\documentclass[letterpaper]{article}

\newtheorem{THEOREM}{Theorem}[section]
\newenvironment{theorem}{\begin{THEOREM} \hspace{-.85em} {\bf :} }%
                        {\end{THEOREM}}
\newtheorem{LEMMA}[THEOREM]{Lemma}
\newenvironment{lemma}{\begin{LEMMA} \hspace{-.85em} {\bf :} }%
                      {\end{LEMMA}}
\newtheorem{COROLLARY}[THEOREM]{Corollary}
\newenvironment{corollary}{\begin{COROLLARY} \hspace{-.85em} {\bf :} }%
                          {\end{COROLLARY}}
\newtheorem{PROPOSITION}[THEOREM]{Proposition}
\newenvironment{proposition}{\begin{PROPOSITION} \hspace{-.85em} {\bf :} }%
                            {\end{PROPOSITION}}
\newtheorem{DEFINITION}[THEOREM]{Definition}
\newenvironment{definition}{\begin{DEFINITION} \hspace{-.85em} {\bf :} \rm}%
                            {\end{DEFINITION}}
\newtheorem{CLAIM}[THEOREM]{Claim}
\newenvironment{claim}{\begin{CLAIM} \hspace{-.85em} {\bf :} \rm}%
                            {\end{CLAIM}}
\newtheorem{EXAMPLE}[THEOREM]{Example}
\newenvironment{example}{\begin{EXAMPLE} \hspace{-.85em} {\bf :} \rm}%
                            {\end{EXAMPLE}}
\newtheorem{REMARK}[THEOREM]{Remark}
\newenvironment{remark}{\begin{REMARK} \hspace{-.85em} {\bf :} \rm}%
                            {\end{REMARK}}

\newcommand{\thm}{\begin{theorem}}
\newcommand{\lem}{\begin{lemma}}
\newcommand{\pro}{\begin{proposition}}
\newcommand{\dfn}{\begin{definition}}
\newcommand{\rem}{\begin{remark}}
\newcommand{\xam}{\begin{example}}
\newcommand{\cor}{\begin{corollary}}
\newcommand{\prf}{\noindent{\bf Proof:} }
\newcommand{\ethm}{\end{theorem}}
\newcommand{\elem}{\end{lemma}}
\newcommand{\epro}{\end{proposition}}
\newcommand{\edfn}{\bbox\end{definition}}
\newcommand{\erem}{\bbox\end{remark}}
\newcommand{\exam}{\bbox\end{example}}
\newcommand{\ecor}{\end{corollary}}
\newcommand{\eprf}{\bbox\vspace{0.1in}}
\newcommand{\beqn}{\begin{equation}}
\newcommand{\eeqn}{\end{equation}}

\newcommand{\bbox}{\vrule height7pt width4pt depth1pt}

\newcommand{\clm}{\begin{claim}}
\newcommand{\eclm}{\end{claim}}

\newcommand{\sat}{\models}

\newcommand{\union}{\cup}
\newcommand{\inter}{\cap}

\renewcommand{\phi}{\varphi}

\newcommand{\F}{{\cal F}}

\newcommand{\R}{{\cal R}}

\newcommand{\U}{{\cal U}}
\newcommand{\V}{{\cal V}}

\newcommand{\ol}{\setlength{\itemsep}{0pt}\begin{enumerate}}
\newcommand{\eol}{\end{enumerate}\setlength{\itemsep}{-\parsep}}
\newcommand{\ul}{\setlength{\itemsep}{0pt}\begin{itemize}}
\newcommand{\dl}{\setlength{\itemsep}{0pt}\begin{description}}
\newcommand{\edl}{\end{description}\setlength{\itemsep}{-\parsep}}
\newcommand{\eul}{\end{itemize}\setlength{\itemsep}{-\parsep}}

\newcommand{\BS}{B^{\scriptscriptstyle \cS}}

\newcommand{\commentout}[1]{}

\newcommand{\bi}{\begin{itemize}}
\newcommand{\ei}{\end{itemize}}
\newcommand{\be}{\begin{enumerate}}
\newcommand{\ee}{\end{enumerate}}

 \newcommand{\fullv}[1]{#1}
 \newcommand{\shortv}[1]{\commentout{#1}}
 \fullv{\usepackage{chicagor}}
 \usepackage{times}
\usepackage{times}
\usepackage{helvet}
\usepackage{courier}
\usepackage{graphicx}

\renewcommand{\S}{{\cal S}}
\newcommand{\ST}{\mbox{{\it ST}}}
\newcommand{\BT}{\mbox{{\it BT}}}
\newcommand{\BH}{\mbox{{\it BH}}}
\newcommand{\SH}{\mbox{{\it SH}}}
\renewcommand{\BS}{\mbox{{\it BS}}}
\newcommand{\RT}{\mbox{{\it RT}}}

\newcommand{\NOT}{\mathit{NOT}}
\newcommand{\TWO}{\mathit{TWO}}
\newcommand{\NW}{\mathit{N}}
\newcommand{\PN}{\mathit{PN}}
\newcommand{\VS}{\mathit{VS}}
\shortv{\renewcommand{\citeyear}{\shortcite}}
\newcommand{\WIN}{\mbox{{\em WIN}}}

\begin{document}
\title{Appropriate Causal Models and the Stability of Causation}
\author{
Joseph Y. Halpern\thanks{Supported in part by NSF grants 
IIS-0911036 and  CCF-1214844, AFOSR grant FA9550-08-1-0438 and
by the DoD 
Multidisciplinary University Research Initiative (MURI) program
administered by AFOSR under grant FA9550-12-1-0040.  
Thanks to Sander Beckers, Isabelle Drouet, Chris 
Hitchcock, and Jonathan Livengood for interesting discussions and useful
comments.  I also thank Isabelle and Jonathan for particularly
careful readings of the paper, which uncovered many typos and
problems.  Finally, I think Thomas Blanchard 
for pointing out a serious problem in an earlier version of
Theorem~\ref{pro:stability}.  A preliminary version of this paper
appears in the {\em Proceedings of the Fourteenth International
Conference on  Principles of Knowledge Representation and Reasoning (KR
2014)}, 2014.}
\\
Cornell University\\
halpern@cs.cornell.edu
}
\maketitle
%
%
%

\begin{abstract}
\emph{Causal models} defined in terms of
structural equations have proved to be quite a powerful way of
representing knowledge regarding causality.  However, a number of
authors have given examples that seem to show that the Halpern-Pearl (HP)
definition of causality \cite{HP01b} gives intuitively unreasonable
answers.  
Here it is shown that, for each of these examples, we can give two stories 
consistent with the description in the example, such that intuitions
regarding causality are quite different for each story.  
By adding additional variables, we can
disambiguate the stories.  Moreover, in the resulting causal models, the
HP definition of causality gives the intuitively correct answer.  
It is also shown that, by adding extra variables, a
modification to the original HP definition made to deal with an example
of Hopkins and Pearl \citeyear{HopkinsP02} may not be necessary.
Given how much can be done by adding extra variables, there might be  a
concern that the notion of causality is somewhat unstable.  Can adding extra
variables in a ``conservative'' way (i.e., maintaining all the relations
between the variables in the original model) cause the 
answer to the question ``Is $X=x$ a  cause of $Y=y$?'' to alternate
between ``yes'' and ``no''?  
It is shown that we can have such alternation infinitely often,
but if we take normality into consideration, we cannot.  Indeed, under
appropriate normality assumptions.
Adding an extra variable can change the answer from ``yes' to ``no'',
but after that, it cannot change back to ``yes''.
\end{abstract}

\section{Introduction}
Causal models defined in terms of
structural equations have proved to be quite a powerful way of
representing knowledge regarding causality.   For example, they have
been used to find causes of errors in software 
\fullv{\shortcite{BBCOT12}}
\shortv{\cite{BBCOT12}}  
 and have
been shown to be useful in predicting human attributions of
responsibility \cite{GL10,LGZ13}.
However, a number of
authors \fullv{\shortcite{Glymouretal10,Hall07,Liv13,Spohn12,Weslake11}}
\shortv{\cite{Glymouretal10,Liv13,Spohn12,Weslake11}}
have given examples that seem to show that the Halpern-Pearl (HP)
definition of causality \cite{HP01b} gives intuitively unreasonable
answers.  One contribution of this paper is to show that these
``problematic'' examples
can be dealt with in a relatively uniform way, by being a little
more careful about the choice of causal model.

The need to choose the causal model carefully has been pointed out frequently
\cite{BS13,Hall07,HP01b,HH10,hitchcock:99,Hitchcock07}.  
A causal model is
characterized by the choice of variables, the equations relating them,
and which variables we choose to make exogenous and endogenous (roughly
speaking, which are the variables we choose to take as given and which
we consider to be modifiable).
Different choices of causal model for a given situation can lead to
different conclusions regarding causality. 
The choices are, to some extent, subjective.   
While some suggestions have been made for good rules of thumb for  
choosing random variables (e.g., in \cite{HH10}), they are certainly not
definitive. Moreover, the choice of variables may also depend in part on
the variables that the modeler is aware of.  

In this paper, I consider the choice of representation in
more detail in \fullv{five} \shortv{four} examples.
I show that in all these examples, 
the model originally considered 
(which I call the ``naive'' model) 
does not
correctly model all the relevant features of the situation.  
I argue that we can see this because, in all these cases,  there is another
story that can be told, also consistent with the naive model,
for which we have quite different intuitions
regarding causality.  This suggests that a more detailed model is needed
to disambiguate the stories.  In the first four cases, what turns
out to arguably be the best way to do the disambiguation is to add
(quite well motivated) extra variables, which, roughly speaking, capture
the mechanism of causality.  In the final example, what turns out to be
most relevant is the decision as to which variables to make exogenous.
Once we model things more carefully, the HP approach gives
the expected answer in all cases.

As already observed by Halpern and Hitchcock \citeyear{HH11}, adding
extra variables also lets us deal with two other concerns that resulted in
changes to the original HP definition.
In Section~\ref{sec:Hopkins}, I consider an example due to Hopkins and Pearl
\citeyear{HopkinsP02} that motivated one of the changes.  After
showing 
how this example can be dealt with by adding an extra variable in a
natural way (without modifying the original HP definition), I show that
this approach generalizes: we can always add extra variables so as to
get a model where the original HP definition can be used.
\fullv{In Section~\ref{sec:normality},}
\shortv{In the full paper,} 
I discuss an example due to Hiddleston
\citeyear{Hiddleston05} that motivated the addition of normality
considerations to the basic HP framework (see Section~\ref{sec:review}).
Again, adding an extra variable deals with this example.

All these examples show that adding extra variables
can result in a cause 
becoming a non-cause.  Can adding variables also
result in a non-cause becoming a cause?  
Of course, without constraints, this can easily happen.  Adding extra
variables can fundamentally change the model.  
Indeed, even if we insist that 
variables are added in a conservative way (so as to maintain all the relations
between the variables in the original model), $X=x$ can alternate
 infinitely often between being a cause of $Y=y$ and not being a cause.  
But, in a precise sense,  this requires the new variables we add to
take on abnormal values. 
Once we talk normality into consideration, this cannot happen.  
If $X=x$ is not a cause of $Y=y$, then adding extra variables to the
model cannot make $X=x$ a cause of $Y=y$.

The rest of this paper is organized as follows.  In the next section, I
review the HP definition (and the original definition) and its extension
to deal with normality, as discussed in \cite{HH11}.  I
discuss the five examples in Section~\ref{sec:examples}.  In
Section~\ref{sec:Hopkins}, 
I discuss how adding extra variables can deal with the Hopkins-Pearl
example and, more generally, can obviate the need to modify the original
HP definition.  
\fullv{
In Section~\ref{sec:normality}, I discuss the extent to 
which adding extra variables can avoid the need to taking normality into
account.} 
In Section~\ref{sec:stability}, I discuss issues of stability.
I conclude in
Section~\ref{sec:conclusions} with some discussion of the implications of
these results. 
\shortv{Details of some proofs are left to the full paper
(available at www.cs.cornell.edu/home/halpern/papers/causalmodeling.pdf).}

\section{Review}\label{sec:review}

In this section, I briefly review the definitions of causal structures,
the HP definition(s) of causality, and the extension that takes into
account normality given by Halpern and Hitchcock.  
The exposition is largely taken from \cite{Hal39}.
The reader is encouraged to consult \cite{HP01b}, and
\cite{HH11} for more details and intuition.

\subsection{Causal models}
The HP approach assumes that the world is described in terms of random
variables and their values.  
Some random variables may have a causal influence on others. This
influence is modeled by a set of {\em structural equations}.
It is conceptually useful to split the random variables into two
sets: the {\em exogenous\/} variables, whose values are
determined by 
factors outside the model, and the
{\em endogenous\/} variables, whose values are ultimately determined by
the exogenous variables.  
For example, in a voting scenario, we could have endogenous variables
that describe what the voters actually do (i.e., which candidate they
vote for), exogenous variables 
that describe the factors
that determine how the voters vote, and a
variable describing the outcome (who wins).  The structural equations
describe how the outcome is determined \fullv{(majority rules; a candidate
wins if $A$ and at least two of $B$, $C$, $D$, and $E$ vote for him;
etc.).}
\shortv{(e.g., majority rules).}

Formally, a \emph{causal model} $M$
is a pair $(\S,\F)$, where $\S$ is a \emph{signature}, which explicitly
lists the endogenous and exogenous variables  and characterizes
their possible values, and $\F$ defines a set of \emph{modifiable
structural equations}, relating the values of the variables.  
A signature $\S$ is a tuple $(\U,\V,\R)$, where $\U$ is a set of
exogenous variables, $\V$ is a set 
of endogenous variables, and $\R$ associates with every variable $Y \in 
\U \union \V$ a nonempty set $\R(Y)$ of possible values for 
$Y$ (that is, the set of values over which $Y$ {\em ranges}).  
For simplicity, I assume here that $\V$ is finite, as is $\R(Y)$ for
every endogenous variable $Y \in \V$.
$\F$ associates with each endogenous variable $X \in \V$ a
function denoted $F_X$ such that $F_X: (\times_{U \in \U} \R(U))
\times (\times_{Y \in \V - \{X\}} \R(Y)) \rightarrow \R(X)$.
This mathematical notation just makes precise the fact that 
$F_X$ determines the value of $X$,
given the values of all the other variables in $\U \union \V$.
If there is one exogenous variable $U$ and three endogenous
variables, $X$, $Y$, and $Z$, then $F_X$ defines the values of $X$ in
terms of the values of $Y$, $Z$, and $U$.  For example, we might have 
$F_X(u,y,z) = u+y$, which is usually written as
$X = U+Y$.%
\footnote{The fact that $X$ is assigned  $U+Y$ (i.e., the value
of $X$ is the sum of the values of $U$ and $Y$) does not imply
that $Y$ is assigned $X-U$; that is, $F_Y(U,X,Z) = X-U$ does not
necessarily hold.}  Thus, if $Y = 3$ and $U = 2$, then
$X=5$, regardless of how $Z$ is set.  

The structural equations define what happens in the presence of external
interventions. 
Setting the value of some variable $X$ to $x$ in a causal
model $M = (\S,\F)$ results in a new causal model, denoted $M_{X
= x}$, which is identical to $M$, except that the
equation for $X$ in $\F$ is replaced by $X = x$.

Following \cite{HP01b}, I restrict attention here to what are called {\em
recursive\/} (or {\em acyclic\/}) models.  This is the special case
where there is some total ordering $\prec$ of the endogenous variables
(the ones in $\V$) 
such that if $X \prec Y$, then $X$ is independent of $Y$, 
that is, $F_X(\ldots, y, \ldots) = F_X(\ldots, y', \ldots)$ for all $y, y' \in
\R(Y)$.  Intuitively, if a theory is recursive, there is no
feedback.  If $X \prec Y$, then the value of $X$ may affect the value of
$Y$, but the value of $Y$ cannot affect the value of $X$.
It should be clear that if $M$ is an acyclic  causal model,
then given a \emph{context}, that is, a setting $\vec{u}$ for the
exogenous variables in $\U$, there is a unique solution for all the
equations.  We simply solve for the variables in the order given by
$\prec$. The value of the variables that come first in the order, that
is, the variables $X$ such that there is no variable $Y$ such that $
Y\prec X$, depend only on the exogenous variables, so their value is
immediately determined by the values of the exogenous variables.  
The values of variables later in the order can be determined once we have
determined the values of all the variables earlier in the order.

\subsection{A language for reasoning about causality}
To define causality carefully, it is useful to have a language to reason
about causality.
Given a signature $\S = (\U,\V,\R)$, a \emph{primitive event} is a
formula of the form $X = x$, for  $X \in \V$ and $x \in \R(X)$.  
A {\em causal formula (over $\S$)\/} is one of the form
$[Y_1 \gets y_1, \ldots, Y_k \gets y_k] \phi$,
where
\fullv{
\begin{itemize}
\item}
$\phi$ is a Boolean
combination of primitive events,
\fullv{\item} $Y_1, \ldots, Y_k$ are distinct variables in $\V$, and
\fullv{\item} $y_i \in \R(Y_i)$.
\fullv{\end{itemize}}
Such a formula is
abbreviated
as $[\vec{Y} \gets \vec{y}]\phi$.
The special
case where $k=0$
is abbreviated as
$\phi$.
Intuitively,
$[Y_1 \gets y_1, \ldots, Y_k \gets y_k] \phi$ says that
$\phi$ would hold if
$Y_i$ were set to $y_i$, for $i = 1,\ldots,k$.

A causal formula $\psi$ is true or false in a causal model, given a
context.
As usual, I write $(M,\vec{u}) \sat \psi$  if
the causal formula $\psi$ is true in
causal model $M$ given context $\vec{u}$.
The $\sat$ relation is defined inductively.
$(M,\vec{u}) \sat X = x$ if
the variable $X$ has value $x$
in the
unique (since we are dealing with acyclic models) solution
to
the equations in
$M$ in context $\vec{u}$
(that is, the
unique vector
of values for the exogenous variables that simultaneously satisfies all
equations 
in $M$ 
with the variables in $\U$ set to $\vec{u}$).
The truth of conjunctions and negations is defined in the standard way.
Finally, 
$(M,\vec{u}) \sat [\vec{Y} \gets \vec{y}]\phi$ if 
$(M_{\vec{Y} = \vec{y}},\vec{u}) \sat \phi$.
I write $M \sat \phi$ if $(M,\vec{u}) \sat \phi$ for all contexts~$\vec{u}$.

\subsection{The definition(s) of causality}
The HP definition of causality, like many others, is based on
counterfactuals.  The idea is that $A$ is a cause of $B$ if, if $A$
hadn't occurred (although it did), then $B$ would not have occurred.
But there are many examples showing that this naive definition will not
quite work.  To take just one example, consider the following story, due
to Ned Hall and already discussed in \cite{HP01b}, from where the
following version is taken.
\begin{quote}
Suzy and Billy both pick up rocks
and throw them at  a bottle.
Suzy's rock gets there first, shattering the
bottle.  Since both throws are perfectly accurate, Billy's would have
shattered the bottle
had it not been preempted by Suzy's throw.
\end{quote}
We would like to say that Suzy's throw is a cause of the bottle
shattering, and Billy's is not.  But if Suzy hadn't thrown, Billy's rock
would have hit the bottle and shattered it.  

The HP definition of causality is intended to deal with this example,
and many others.

\dfn\label{actcaus}
$\vec{X} = \vec{x}$ is an {\em actual cause of $\phi$ in
$(M, \vec{u})$ \/} if the following
three conditions hold:
\begin{description}
\item[{\rm AC1.}]\label{ac1} $(M,\vec{u}) \sat (\vec{X} = \vec{x})$ and 
$(M,\vec{u}) \sat \phi$.
\item[{\rm AC2.}]\label{ac2}
There is a partition of $\V$ (the set of endogenous variables) into two
subsets $\vec{Z}$ and $\vec{W}$%
\footnote{I occasionally use the vector notation ($\vec{Z}$, $\vec{W}$, etc.)
to denote a set of variables if the order of the variables matters,
which it does when we consider an assignment such as $\vec{W} \gets \vec{w}$.}
with $\vec{X} \subseteq \vec{Z}$ and a
setting $\vec{x}'$ and $\vec{w}$ of the variables in $\vec{X}$ and
$\vec{W}$, respectively, such that
if $(M,\vec{u}) \sat Z = z$ for 
all $Z \in \vec{Z}$
(i.e., $z$ is the value of the random variable $Z$ in the real world),
then
both of the following conditions hold:
\begin{description}
\item[{\rm (a)}]
$(M,\vec{u}) \sat [\vec{X} \gets \vec{x}',
\vec{W} \gets \vec{w}]\neg \phi$.
\item[{\rm (b)}]
$(M,\vec{u}) \sat [\vec{X} \gets
\vec{x}, \vec{W}' \gets \vec{w}, \vec{Z}' \gets \vec{z}]\phi$ for 
all subsets $\vec{W}'$ of $\vec{W}$ and all subsets $\vec{Z'}$ of
$\vec{Z}$, where I abuse notation and write $\vec{W}' \gets \vec{w}$ to
denote the assignment where the variables in $\vec{W}'$ get the same
values as they would in the assignment $\vec{W} \gets \vec{w}$, and
similarly for $\vec{Z}' \gets \vec{z}$. 
\end{description}
\item[{\rm AC3.}] \label{ac3}
$\vec{X}$ is minimal; no subset of $\vec{X}$ satisfies
conditions AC1 and AC2.
\label{def3.1}  
\end{description}
The tuple $(\vec{W}, \vec{w}, \vec{x}')$ is said to be a
\emph{witness} to the fact that $\vec{X} = \vec{x}$ is a cause of
$\phi$.
\end{definition}

AC1 just says that $\vec{X}=\vec{x}$ cannot
be considered a cause of $\phi$ unless both $\vec{X} = \vec{x}$ and
$\phi$ actually happen.  AC3 is a minimality condition, which ensures
that only those elements of 
the conjunction $\vec{X}=\vec{x}$ that are essential 
\fullv{for changing $\phi$ in AC2(a)} are
considered part of a cause; inessential elements are pruned.
Without AC3, if dropping a lit cigarette is a cause of a fire
then so is dropping the cigarette and sneezing.
\fullv{AC3 serves here to strip ``sneezing''
and other irrelevant, over-specific details
from the cause.}

AC2 is the core of the definition.
We can think of the variables in $\vec{Z}$ as making up the ``causal
path'' from $\vec{X}$ to $\phi$.  Intuitively, changing the value of
some variable in $X$ results in changing the value(s) of some
variable(s) in $\vec{Z}$, which results in the values of some
other variable(s) in $\vec{Z}$ being changed, which finally results in
the value of $\phi$ changing.  The remaining endogenous variables, the
ones in $\vec{W}$, are off to the side, so to speak, but may still have
an indirect effect on what happens.
AC2(a) is essentially the standard
counterfactual definition of causality, but with a twist.  If we 
want to show that $\vec{X} = \vec{x}$ is a cause of $\phi$, we must show
(in part) that if $\vec{X}$ had a different value, then so too would
$\phi$.  However, the effect on $\phi$ of changing the value of the variables
in $\vec{X}$ may not obtain unless we also change  the values
of some of the ``off path'' variables in $\vec{W}$.  Intuitively, 
setting $\vec{W}$ to $\vec{w}$ eliminates
some side effects that may mask the effect of changing the
value of $\vec{X}$.  For example, if Billy and Suzy both throw rocks
at a bottle and hit it simultaneously, shattering it, but one rock 
would have sufficed to shatter the bottle, then to show that Billy's
throw is a cause of the bottle shattering, we consider a setting where
Suzy does not throw.  Then if Billy doesn't throw, the bottle doesn't
shatter, while if he throws it does shatter.
We do require that, although the values of variables on the
causal path (i.e.,  the variables $\vec{Z}$) may be perturbed by
the change to $\vec{W}$, this perturbation has no impact on the value of
$\phi$.  As  I said when defining AC2, if $\vec{u}$ is the actual
context and $(M,\vec{u}) \sat \vec{Z} = \vec{z}$,
then $z$ 
is the value of the 
variable $Z$ in the actual situation.
We capture the fact that the 
perturbation has no impact on the value of $\phi$ by saying that if some
variables $Z$ on the causal path were set to their values in the
context $\vec{u}$, $\phi$ would still be true, as long as $\vec{X} =
\vec{x}$.
Roughly speaking, AC2(b) says that if the variables in $\vec{X}$ are reset
to their original value, then $\phi$ holds, even if only a subset
$\vec{W}'$ of the variables in $\vec{W}$ are set to their values in
$\vec{W}$ and and even if some variables in $\vec{Z}$ are set to
their original values (i.e., the values in $\vec{z}$).
The fact that AC2(b) must hold even if only a subset $\vec{W}'$ of the
variables in 
$\vec{W}$ are set to their values in $\vec{w}$ (so that the variables in
$\vec{W} - \vec{W}'$ essentially act as they do in the real world;
that is, they are allow to vary freely, according to the structural
equations, rather than being set to their values in $\vec{w}$) 
and only a subset of the
variables in $\vec{Z}$ are set to their values in the actual world says
that we must have $\phi$ even if some things happen as they do in the
actual world.  See Sections~\ref{subsec:rt} and \ref{sec:Hopkins} for
further discussion of and intuition for AC2(b).

The original HP paper \cite{HPearl01a} used a weaker version
of AC2(b).  Rather than requiring that
$(M,\vec{u}) \sat [\vec{X} \gets
\vec{x}, \vec{W}' \gets \vec{w}, \vec{Z}' \gets \vec{z}]\phi$ 
for all subsets $\vec{W}'$ of $\vec{W}$,
it was required to
hold only for $\vec{W}$.  That is, the following condition was used
instead of AC2(b).  
\begin{description}
\item[AC2(b$'$)] 
$(M,\vec{u}) \sat [\vec{X} \gets
\vec{x}, \vec{W} \gets \vec{w}, \vec{Z}' \gets \vec{z}]\phi$ for 
all subsets $\vec{Z'}$ of
$\vec{Z}$. 
\end{description}
The change from AC2(b$'$) to AC2(b) may seem rather technical, but it
has some nontrivial consequences.  One of the contributions of this
paper is to examine whether it is necessary; see
Section~\ref{sec:Hopkins} for details.

To deal with other problems in the HP definition, various authors have
added the idea of \emph{normality} to the definition.  This can be done
in a number of ways.  I now briefly sketch one way that this
can be done, following the approach in \cite{HH11}.  
\fullv{(See Section~\ref{sec:normality} for some discussion of the need for normality.)}

Take a \emph{world} (in a model $M$) to be a complete assignment of
values to the endogenous variables in $M$.%
\footnote{In \cite{HH11}, a world is defined as a  complete assignment
of values to the \emph{exogenous} variables, but this is a typo.}
(See the discussion after Corollary~\ref{cor:stability} for why it is
conimportant that a world is an assignment only to 
the endogenous variables, and not all the variables, including the
exogenous variables.)
We assume a partial preorder $\succeq$ 
on worlds, that is, a reflexive transitive relation.%
\footnote{$\succeq$ is not necessarily a partial order; in particular,
it does not necessarily satisfy \emph{antisymmetry} (i.e., $s  \succeq
s'$ and $s' \succeq s$ does not necessarily imply $s=s'$).}  
Intuitively, if $s \succeq s'$, then $s$ is at least as normal, or
typical, as $s'$.   We can use normality in the definition of causality
in two ways.  Say that a world $s$ is a \emph{witness world} for $\vec{X} =
\vec{x}$ being a cause of $\phi$ in $(M,\vec{u})$ if there is a witness 
$(\vec{W}, \vec{w}, \vec{x}')$ to $\vec{X} = \vec{x}$ being a cause of
$\phi$ and $s=s_{\vec{X} = \vec{x}', \vec{W} = \vec{w},\vec{u}}$, where
$s_{\vec{X} = \vec{x}', \vec{W} = \vec{w},\vec{u}}$ 
is the world that results by setting $\vec{X}$ to $\vec{x}'$ and $\vec{W}$
to $\vec{w}$ in context $\vec{u}$.
We can then modify AC2(a) so as to require that we
consider $\vec{X} = \vec{x}$ to be a cause of $\phi$ in $(M,\vec{u})$
only if the witness world $s$ for $\vec{X} = \vec{x}$ being a cause is such that
$s \succeq s_{\vec{u}}$, where $s_{\vec{u}}$ is the world
determined by context $\vec{u}$;
call this modified version AC2(a$^+$).
AC2(a$^+$) says that, in determining causality, we consider only 
possibilities that result from altering atypical features of a world to 
make them more typical, rather than vice versa.  This captures an
observation made by Kahneman and Miller \citeyear{KM86} regarding human
ascriptions of causality.
An \emph{extended causal model} is a causal model together with a
preorder $\succeq$ on worlds.  
Say that \emph{$\vec{X} = \vec{x}$ is a cause of $\phi$ according
  to the extended HP definition in $(M,\vec{u})$}, where $M$ is an
  extended causal model, if $\vec{X} = \vec{x}$ is a cause of
$\phi$ using 
AC2(a$^+$) rather than AC2(a).

A somewhat more refined use of normality is to use it to ``grade''
causes.  Say that $s$ is a \emph{best witness} for 
$\vec{X} = \vec{x}$ being a cause of $\phi$ if $s$ is a witness world for 
$\vec{X} = \vec{x}$ being a cause of $\phi$ and 
there is no other witness world $s'$ 
for $\vec{X} = \vec{x}$ being a cause of $\phi$ 
such that $s' \succ s$.  (Note that
there may be more than one best witness.)  We can then grade candidate
causes according to the normality of their best witnesses (without
requiring that there must be a witness $s$ such that $s \succeq s_{\vec{u}}$).
Experimental evidence suggests that 
people are focusing on the cause with the best witness (according to
their subjective ordering on worlds); 
see, e.g., \cite{CKS08,HK09,KF08}.

\section{The Examples}\label{sec:examples}

In this section, I consider examples due to 
Spohn \citeyear{Spohn12}, \fullv{Weslake \citeyear{Weslake11},} Hall
\citeyear{Hall07}, 
Glymour et al. \citeyear{Glymouretal10}, and Livengood \citeyear{Liv13}.
I go through these examples in turn.  
I set the scene by considering the rock-throwing example mentioned
above.

\subsection{Throwing rocks at bottles}\label{subsec:rt}
\begin{figure}[htb]
{\begin{center}
\setlength{\unitlength}{.18in}
\begin{picture}(8,8)
\put(3,0){\circle*{.2}}
\put(3,8){\circle*{.2}}
\put(0,4){\circle*{.2}}
\put(6,4){\circle*{.2}}
\put(3,8){\vector(3,-4){3}}
\put(3,8){\vector(-3,-4){3}}
\put(0,4){\vector(3,-4){3}}
\put(6,4){\vector(-3,-4){3}}
\put(3.4,-.2){$\BS$}
\put(-1.0,3.8){$\ST$}
\put(6.15,3.8){$\BT$}
\put(3.4,7.8){$U$}
\end{picture}
\end{center}
}
\caption{$M_{\RT}$: the naive rock-throwing model.}\label{fig:rt}
\end{figure}
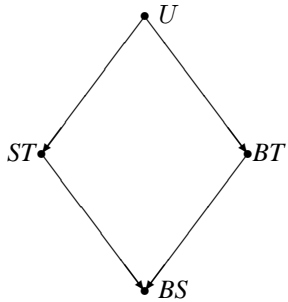

A naive model of the rock-throwing story just has three \emph{binary} random
variables $\ST$, 
$\BT$, and $\BS$ (for ``Suzy throws'', ``Billy throws'', and ``bottle
shatters'').  The fact that the variables are binary means
that they take values in $\{0,1\}$.  The values of $\ST$ and $\BT$ are
determined  by the context; 
the value of $\BS$ given by the equation $\BS= \ST \lor
\BT$: the bottle shatters if Suzy or Billy throws.%
\footnote{Here and elsewhere, I follow the fairly standard
  mathematical convention of eliding the ``and only if'' in
  definitions.  What is intended here is that the bottle shatters
  \emph{if and only if} Suzy or Billy throws.}
Call this model $M_{\RT}$.  
For simplicity, suppose that there is just one exogenous variable.
Let $u$ be the 
context that results in $\ST = \BT = 1$: Suzy and Billy both throw.
$M_{\RT}$ is described in Figure~\ref{fig:rt}. (Although I have included
the exogenous variable here, in later figures exogenous variables are
omitted for ease of presentation.) 

As already pointed out by Halpern and Pearl \citeyear{HP01b}, in $M_{\RT}$
Suzy and Billy play completely symmetric roles.  Not
surprisingly, both $\ST=1$ and $\BT=1$ are causes of $\BS=1$ according
to the HP definition.  Clearly, $M_{\RT}$ cannot be used to distinguish
a situation where Suzy is a cause from one where Billy is a cause.

\begin{figure}[htb]
{\begin{center}
\setlength{\unitlength}{.18in}
\begin{picture}(8,9)
\put(3,0){\circle*{.2}}
\put(0,8){\circle*{.2}}
\put(6,8){\circle*{.2}}
\put(0,4){\circle*{.2}}
\put(6,4){\circle*{.2}}
\put(0,8){\vector(0,-1){4}}
\put(6,8){\vector(0,-1){4}}
\put(0,4){\vector(1,0){6}}
\put(0,4){\vector(3,-4){3}}
\put(6,4){\vector(-3,-4){3}}
\put(3.4,-.2){$\BS$}
\put(-1.0,7.8){$\ST$}
\put(6.15,7.8){$\BT$}
\put(-0.9,3.8){$\SH$}
\put(6.15,3.8){$\BH$}
\end{picture}
\end{center}
}
\caption{$M_{\RT}'$: the better rock-throwing model.}\label{fig:rt1}
\end{figure}
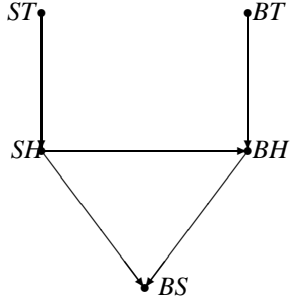

In the story as given, people seem to agree that Suzy's throw is a 
cause and Billy's throw is not, since Suzy's rock hit
the bottle and Billy's did not.  $M_{\RT}$ does not capture this fact.
Following Halpern and Pearl \citeyear{HP01b}, we extend $M_{\RT}$ so that 
it can express the fact that Suzy's rock hit first by adding two more
variables: 
\begin{itemize}
\item $\BH$ for ``Billy's rock hits the (intact) bottle'', with values 0
(it doesn't) and 1 (it does); and
\item $\SH$ for ``Suzy's rock hits the bottle'', again with values 0 and
1.
\end{itemize}
The equations are such that $\SH = \ST$ (Suzy's rock hits the bottle
if Suzy throws), $\BH = \BT \land \neg \SH$ (Billy's rock hits an
intact bottle if Billy throws and Suzy's rock does not hit), and 
$\BS = \SH \lor \BH$ (the bottle shatters if either Suzy's rock or
Billy's rock hit it).  
Now if Suzy and Bill both throw ($\ST=1$ and $\BT=1$), Suzy's rock hits
the bottle ($\SH = 1$), so that Billy's rock does not hit an intact
bottle ($\BH=0$).  Call the resulting model $M_{\RT}'$.
$M_{\RT}'$ is described in Figure~\ref{fig:rt1} (with the exogenous
variable omitted).

In this model, $\BT=1$ is \emph{not} a cause.  For example, if we take
$\vec{Z} = \{\BT, \BH, \BS\}$ in AC2 and set $\ST = 0$, then while it is
the case that $\BS=0$ if $\BT = 0$ and $\BS=1$ if $\BT=1$, it is not the
case that $\BT=1$ if we set $\BH$ to its original value of 0.  
Similar arguments work for all other partitions into $\vec{Z}$
and $\vec{W}$.  The key is to consider whether 
$\BH$ is in $\vec{W}$ or $\vec{Z}$.  If $\BH$ is in $\vec{W}$, then
how we set $\BT$ has no effect on the value $\BS$, so $\BT=1$ cannot
be cause.  And if  $\BH$ is in $\vec{W}$, then we get the same problem
with with AC2(b$^o$) or AC2(b$^u$) as above, 
since it is easy to see that at least one of 
$\SH$ or $\ST$ must in $\vec{W}$, and $\vec{w}$ must be such that
whichever is in $\vec{W}$ is set to 0.    I leave the details to the
reader.  

This example shows that it necessary in AC2(b) to allow some
variables, but not necessarily all, variables in $\vec{Z}$ to be
set to their original values.  For example, if we take $\vec{W} =
\{\ST\}$,  $\vec{Z} = \{\BT,\BH,\BS,\SH\}$, and $\vec{w} = 0$, to show
that $\BT = 1$ is not a cause of $\BS = 1$, we must set $\BH = 0$, its
original value, but we do \emph{not} want to set $\SH = 1$. 
Setting $\BH$ to 0 captures the intuition
that Billy's throw is not a cause because, in the actual world, his rock
did not hit the bottle ($\BH = 0$).  By AC2(b), to establish $\BT=1$
as a cause of $\BS=1$, setting $\BT$ to 1 would have to force $\BS=1$
even if $\BH=0$, 
which is not the case.   

\subsection{Spohn's example}
The next example is due to Spohn \citeyear{Spohn12}.

\xam\label{xam:spohn}
There are four endogenous binary variables, $A$, $B$, $C$, and $S$,
taking values 1 (on) and 0 (off). Intuitively, $A$ and $B$ are 
supposed to be alternative causes of $C$, and $S$ acts as a switch. If $S=0$,
the causal route from $A$ to $C$ is active and that from $B$ to $C$ is
dead; and if $S=1$, the causal route from $A$ to $C$ is dead and the one
from $B$ to $C$ is 
active. There are no causal
relations between $A$, $B$, and $S$; their values are determined by the
context.
The equation for $C$ is  $C= (\neg S \land A) \lor (S \land
B)$.

Suppose that the context is such that $A=B=S=1$, so $C=1$.  
The HP definition yields
 $B=1$ and $S=1$ as causes of $C=1$, as we would hope.  But,
unfortunately, it also yields $A=1$ as a cause of $C=1$.  The argument
is that in the contingency where $S$ is set to 0, if $A=0$, then
$C=0$, 
while if $A=1$, then $C=1$.  This does not seem so
reasonable. Intuitively, if $S=1$, then the value of $A$ seems
irrelevant to the outcome. 
Considerations of normality do not help here; all worlds seem to
be equally normal.

But now consider a slightly different story.   This time,
we view $B$ as the switch, rather than $S$.  If $B=1$, then $C=1$ if
either $A=1$ or $S=1$; if $B=0$, then $C=1$ only if $A=1$ and $S=0$.
That is, $C = (B \land (A \lor S)) \lor (\neg B \land A \land \neg S)$.
Although this is perhaps not as natural a story as the original, such a
switch is surely implementable.  In any case, a little playing with
propositional logic shows that, in this story, $C$ satisfies exactly the
same equation as before: $(\neg S \land A) \lor (S \land
B)$ is equivalent to $(B \land (A \lor S)) \lor (\neg B \land A \land \neg S)$.
The key point is that, unlike the first story,
in the second story, it seems to me quite
reasonable to say that $A=1$ is a cause of $C=1$ (as are $S=1$ and
$B=1$).  Having $A=1$ is necessary for the first ``mechanism'' to work.

Given that we have different causal intuitions for the stories, we should
model them differently.  One way to distinguish them is to add two
more endogenous random variables, say $D$ and $E$, that describe the
ways that $C$ could be 1.  
In Spohn's original story, we would have the equation  $D= \neg S \land
A$, $E= S \land B$, and 
$C = D \lor E$. In this  model,
since $D=0$ in the actual context, it is not hard to see that $A=1$ is
\emph{not} a cause of $C=1$, while $B=1$ and 
$S=1$ are, as they should be.  Thus, in this model, we correctly capture
our intuitions for the story.

To capture the second story, we can add variables $D'$ and
$E'$ such that
$D'= B \land (A \lor S)$,  $E'= \neg B \land A
\land S$, and $C = D' \lor E'$.  In this model, it is not
hard to see that all of $A=1$, $B=1$, and $S=1$ are causes of $C=1$.

This approach of adding extra variables leads to an obvious question:
What is the role of these variables? I view $D$ and $E$ (resp., $D'$ and
$E'$) as ``structuring'' variables, that help an agent ``structure'' a
causal story.  Consider Spohn's original story.
We can certainly design a circuit where there is a source of power at
$A$ and $B$, a physical switch at $S$, and a bulb at $C$ that turns on
($C=1$) if either there is a battery at $A$ ($A=1$) and the switch is
turned left ($S=0$) or there is battery at $B$ ($B=1$) and the switch is turned
right ($S=1$).  In this physical setup, there is no analogue of $D$ and
$E$.  Nevertheless, to the extent that we view the models as a modeler's
description of what is going, a modeler could usefully introduce $D$ and
$E$ to describe the conditions under which $C=1$, and to disambiguate
this model from one where, conceptually, we might want to think of other
ways that $C$ could be 1 (as in the story with $D'$ and $E'$).

Note that we do not want to think of $D$ as being \emph{defined} 
to take the value 1 if $A=1$ and $S=0$.  For then we could not intervene
to set $D=0$ if $A=1$ and $S=0$. Adding a
variable to the model commits us to be able to intervene on it.%
\footnote{I thank Chris Hitchcock for stressing this point.}  
In the real world, setting $D$ to 0 despite having $A=1$ and $S=0$ might
correspond to the connection being faulty when the switch is turned
left.  Indeed, since the equation for $C$ is the same in both stories,
it is only at the level of interventions that the difference
between the two stories becomes meaningful.  
\exam

\fullv{
\subsection{Weslake's example}
The next example is due to Weslake \citeyear[Example 10]{Weslake11}.

\xam  A lamp $L$ is controlled by three
switches, $A$, $B$, and $C$, each of which has three possible positions,
$-1$, $0$, and $1$.  The lamp switches on iff two or more of the
switches are in same position.  Thus, $L=1$ iff $
(A=B) \lor (B=C) \lor (A=C).$
Suppose that, in
the actual context, $A=1$, $B=-1$, and $C=-1$.  Intuition
suggests that while $B=-1$ and $C=-1$ should be causes of $L=1$, $A=1$
should not be; since the setting of $A$ does not match that of either $B$
or $C$, it has no causal impact on the outcome.
The HP definition indeed declares $B=-1$ and $C=-1$ to be
causes; unfortunately, it also declares $A=1$ to be a cause.  For in the
contingency where $B=1$ and $C=-1$, if $A=1$ then $L=1$, while if $A=0$ then
$L=0$.  Adding defaults to the picture does not solve the
problem.  

Just as in the Spohn example, we can tell another story where the
observed variables have the same values, and are connected by the same
structural equations.  
Now suppose that $L=1$ iff 
either (a) none of $A$, $B$, or $C$ is in position $-1$,  (b) none of $A$,
$B$, or $C$ is in position 0, or (c) none of $A$, $B$, or $C$ is in position 1.
It is easy to see that the equations for $L$ are 
literally the same as in the original example.
But now it seems more reasonable 
to say that $A=1$ \emph{is} a cause of $L=1$.  Certainly $A=1$ causes
$L=1$ as a result of no values being 0; had $A$ been 0, then the lamp
would still 
have been on, but now it would be as a result of no values being $-1$.
Considering the contingency where $B=1$ and $C=-1$ ``uncovers'' the
causal impact of $A$.  

Again, we can capture the distinction between the two stories by adding more
variables.  For the second story, we can add the variables $\NOT(-1)$,
$\NOT(0)$, and $\NOT(1)$, where $\NOT(i)$ is 1 iff none of $A$, $B$, or
$C$ are $i$. Then $L = \NOT(-1) \lor \NOT(0) \lor
\NOT(1)$.  Now the HP definition makes $A=1$ a cause of $L=1$ (as well
as $B=-1$ and $C=-1$).  For Weslake's original story
we can add the variables $\TWO(-1)$, $\TWO(0)$, and $\TWO(1)$, where
$\TWO(i)=1$ 
iff at least two of $A$, $B$, and $C$ are $i$, and take $L = \TWO(-1) \lor
\TWO(0) \lor TWO(1)$.  Now the HP definition does not make $A=1$ a cause of
$L=1$ (although, of course $B=1$ and $C=1$ continue to be causes).
%

Once again, I think of the variables $\NOT(-1)$,
$\NOT(0)$, and $\NOT(1)$ (resp., $\TWO(-1)$, $\TWO(0)$, and $\TWO(1)$)
as ``structuring'' variables, that help the modeler distinguish the two
scenarios.  They are conceptually meaningful even if they don't have a
physical analogue.
\exam
}

\subsection{Hall's example}
Hall's \citeyear{Hall07} gives an example that's meant to illustrate
how a bad choice of variables leads to unreasonable answers.  I repeat
it here because, although I agree with his main point (that, indeed,
is one of the main points of this paper!), I disagree with one of his
conclusions.  What I present is actually a slightly simplified version
of his example that retains all the necessary features.

Consider a model $M$ with four endogenous variables, $A$, $B$, $D$, and
$E$.  The values of $A$ and $D$ are 
determined by the context.  The values of $B$ and $E$ are given by
the equations $B = A$ and $E=D$.%
\footnote{Hall \citeyear{Hall07} also has variables $C$ and $F$ such
  that $C = B$ and 
  $F=E$; adding them does not affect any of the discussion
  here (or in Hall's paper).}
Suppose that the context $u$ is such that $A=D=1$.  Then clearly, in
context $(M,u)$, $A=1$ is
a cause of $B=1$ and not a cause of $E=1$, while $D=1$ is a cause of
$E=1$ and not of $B=1$.  The problem comes if we replace $A$ in the
model by $X$, where intuitively, $X=1$  iff the context would have
been such that $A$ and $D$ agree (i.e., $X=1$ in the context where 
$A=D=1$ or $A=D=0$).  Now we can recover the value of $A$ from
that of $D$ and $X$; it is easy to see that $A=1$ iff $X=D=1$ or
$X=D=0$.  Thus, we can 
rewrite the equation for $B$ by taking $B=1$ iff $X=D=1$ or $X=D=0$.
Formally, consider a model $M'$ with endogenous variables 
$X$, $B$, $D$, and $E$; the context determines the value of $X$ and
$D$; the equation for $B$ is that given above; and we still have the
equation $E=D$.  Now let $u$ be the context where $X=D=1$.
In $(M',u)$, it is still
the case that $D=1$ is a cause of $E=1$, but now $D=1$ is also a
cause of $B=1$.  

Hall \citeyear{Hall07} says ``This result is plainly silly, and
doesn't look any less silly if you insist that causal claims must
always be relativized to a model.''  I disagree.  To be more precise,
I would argue that Hall has in mind a particular picture of the world,
that captured by model $M$.  Of course, if that is the ``right''
picture of the world, the conclusion that $D=1$ is a cause of $B=1$ is
indeed plainly silly.  But consider the following two stories.  We are
trying to determine the preferences of two people, Betty and Edward,
in an election.  $B=1$ if Betty is recorded as preferring the Democrats
and $B=0$ if Betty is recorded as preferring the Republicans, and
similarly for $E$.   
In the first story, we send Alice to talk to Betty and David to talk to
find out their preferences (both are assumed to be truthful and good at
finding things out).  When Alice reports that Betty prefers the
Democrats ($A=1$) then Betty is reported as preferring the Democrats
($B=1$); similarly for David and Edward.  Clearly, in this story
(which is modeled by $M$) $D=1$ causes $E=1$, but not $B=1$.  

But now suppose instead of sending Alice to talk to Betty, Xavier is
sent to talk to Carol, who
knows only whether Betty and Edward have the same preferences. Carol
tells Xavier that they indeed have the same preferences ($X=1$).  Upon
hearing that $X=D=1$, the vote tabulator correctly concludes that
$B=1$.  This story is modeled by $M'$.  But in this case it strikes me
as perfectly reasonable that $D=1$ should be a cause of $B=1$.  This
is true despite that fact that if we had included the variable $A$ in
$M'$, it would have been the case that $A=B=1$.

\subsection{Glymour et al.'s example}

The next example is due to Glymour et al. \citeyear{Glymouretal10}.  
\xam A ranch has five individuals: $a_1, \ldots, a_5$.  They
have to vote on two possible outcomes: staying at the campfire
($O=0$) or going on a round-up ($O=1$).  Let $A_i$ be the random
variable denoting $a_i$'s vote, so $A_i = j$ if $a_i$ votes for outcome
$j$.  There is a complicated rule for deciding on the outcome.  If $a_1$
and $a_2$ agree (i.e., if $A_1 = A_2$), then that is the outcome.  If
$a_2, \ldots, a_5$ agree, and $a_1$ votes differently, then the outcome
is given by $a_1$'s vote (i.e., $O = A_1$).  Otherwise, majority rules.
In the actual situation, $A_1 = A_2 = 1$ and $A_3 = A_4 = A_5 = 0$, so
by the first mechanism, $O=1$.  The question is what were the causes of
$O=1$.

Using the naive causal model with just the variables $A_1, \ldots, A_5,
O$, and the obvious equations describing $O$ in terms of $A_1, \ldots,
A_5$, it is almost immediate that $A_1 = 1$ is a cause of $O=1$.  Changing
$A_1$ to 0 results in $O=0$.  Somewhat surprisingly, in this naive model,
$A_2 = 1$, $A_3 = 0$, $A_4 =0$, and $A_5 = 0$ are also causes.%
\footnote{Glymour
et al.~point out that $A_1 = 1$, $A_3= 0$, $A_4 = 0$, and $A_5 = 0$ are
causes;
they do not mention that $A_2 = 1$ is also a cause.} To see that $A_2 = 1$
is a cause, consider the contingency where $A_3 = 1$.  Now if $A_2 = 0$,
then $O=0$ (majority rules); if $A_2 = 1$, then $O=1$, since $A_1 = A_2
=1$, and $O=1$ even if $A_3$ is set back to its original value of 0.
To see that $A_3=0$ is a cause, consider the contingency where $A_2 =
0$, so that all voters but $a_1$ vote for 0 (staying at the campsite).
If $A_3 =1$, then $O=0$ (majority rules).  If $A_3 = 0$, then 
$O=1$, by the second mechanism ($a_1$ is the only vote for $0$), while if
$A_2$ is set to its original value of 1, then we still have $O=1$, now
by the first mechanism.

But all this talk of mechanisms (which is also implicit in Glymour et
al. \citeyear{Glymouretal10}; in footnote 11, they say that setting $A_2$
back to its original value of 1 ``brings out the original result, but in
a different way'') suggests that the mechanism should be part of the
model.  There are several ways of doing this.  One is to add three new
variables, call them $M_1$, $M_2$, and $M_3$.  These variables have
values in $\{0,1,2\}$, where $M_j = 0$ if mechanism $j$ is active and
suggests an outcome 0, $M_j = 1$ if mechanism $j$ is active and suggests
an outcome of 1, and $M_j = 2$ if mechanism $j$ is not active.  (We
actually don't need the value $M_3 = 2$; mechanism 3 is always active,
because there is always a majority with 5 voters, all of whom must vote.)
Note that at most one of the first two mechanisms can be active.  We
have obvious equations linking the value of $M_1$, $M_2$, and $M_3$ to
the values of $A_1, \ldots, A_5$.  

Now the value of $O$ just depends on
the values of $M_1$, $M_2$, and $M_3$: if $M_1 \ne 2$, then $O = M_1$;
if $M_2 \ne 2$, then $O = M_2$, and if $M_1 = M_2 = 2$, then $O = M_3$.
It is easy to see that in this model, if $A_1 = A_2 = 1$ and $A_3
=  A_4 = A_5 = 0$, then none of $A_3 = 0$, $A_4=0$, and $A_5 =
0$ is a cause.  $A_1 = 1$ is cause, as we would expect, as are $A_2 =
1$ and $M_2 = 1$.  This seems reasonable: the second mechanism
was the one that led to the outcome, and it required $A_1 = A_2 = 1$.

Now suppose that we change the description of the voting rule.  We take
$O=1$ if one of the following two mechanisms applies:
\begin{itemize}
\item $A_1=1$ and it is not the case that both $A_2=0$ and exactly one of
$A_3$, $A_4$, and $A_5$ is 1.
\item $A_1=0$, $A_2=1$, and exactly two of $A_3$, $A_4$, and $A_5$ are
1.
\end{itemize}
It is not hard to check that, although the description is different,
$O$ satisfies the same equation in both stories.  But now it does not
seem so unreasonable that $A_2=1$, $A_3=0$, $A_4=0$, and $A_5 = 0$ are
causes of $O=1$.  And indeed, if we construct
a model in terms of these two mechanisms (i.e., add variables $M_1'$
and $M_2'$ that correspond to these two mechanisms), then it is not hard
to see that $A_1 =1$, $A_2=1$, $A_3=0$, $A_4=0$, and $A_5 = 0$ are all
causes.
%

Here the role of the structuring variables $M_1$, $M_2$, and $M_3$
(resp. $M_1'$ and $M_2'$) as descriptors of the mechanism being invoked
seems particularly clear.  For example, setting $M_1 = 2$ says that the
first mechanism will not be applied, even if $A_1 = A_2$; setting $M_1 =
1$ says that we act as if both $a_1$ and $a_2$ voted in favor, even if
that is not the case. 
\exam

\subsection{Livengood's voting examples}
As Livengood \citeyear{Liv13} points out, voting can
lead to some apparently unreasonable causal outcomes (at least, if we
model things naively).  He first considers Jack and Jill, who live in an
overwhelmingly Republican district.  As expected, the Republican
candidate wins with an overwhelming majority.  Jill would normally have
voted Democrat, but did not vote because she was disgusted by the
process.  Jack would normally have voted Republican, but did not vote
because he (correctly) assumed that his vote would not affect the
outcome.    In the naive model, both Jack and Jill are causes of the
Republican victory.  
For if enough of the people who voted Republican had switched to voting
Democrat, then if Jack (or Jill) had voted Democrat, the Democrat
would have won, while he would not have won had they abstained.
Notice that, in this argument, Jack and Jill are
treated the same way; their preferences make no difference.

We can easily construct a model that takes these preferences into
account.  One way to do so is to assume that their preferences are so
strong that we may as well take them for granted.  Thus, the
preferences become exogenous; the only endogenous variables are whether
or not they vote.
In this case, Jack's not voting is not a cause of the outcome, but
Jill's not voting is.  

More generally, with this approach, a voter whose preference is made
exogenous and is a strong supporter of the victor does not count as a
cause of victory.  This does not seem so unreasonable.  
After all, in an analysis of a close political victory in
Congress, when an analyst talks about the cause(s) of victory, she points
to the swing voters who voted one way or the other, not the voters that
were taken to be staunch supporters of one particular side.

That said, making a variable exogenous seems like a somewhat
draconian solution to the problem.  It also does not allow us to take into
account smaller gradations in depth of feeling.  At what point should a
preference switch from being endogenous to exogenous?
We can achieve the same effect in an arguably more natural way by
using normality considerations.  In the case of Jack and Jill,
we can take voting for a Democrat to be highly abnormal for Jack, and 
voting for a Republican to be highly abnormal for Jill.  To show that
either Jack (resp., Jill) is a cause of the victory, we need to consider
a contingency where Jack (resp., Jill) votes for the Democratic candidate.  This
would be a change to a highly abnormal world in the case of Jack, but to
a more normal world in the case of Jill.  Thus, if we use normality as a
criterion for determining causality, Jill would count as a cause, but
Jack would not.  If we use normality as a way of grading causes,
Jack and Jill would
still both count as causes for the victory, but Jill would be a much
better cause.  More generally, the more normal it would be for someone
to vote Democrat, the better a cause that voter would be.   The use of
normality here allows for a more nuanced gradation of cause than the
rather blunt approach of either making a variable exogenous or endogenous.

Now, following Livengood \citeyear{Liv13}, consider
 a vote where everyone can either vote for one
of three candidates.  Suppose that the actual vote is 17--2--0 (i.e.,
17 vote for candidate $A$, 2 for candidate $B$, and none for candidate
$C$).   Then not only is every vote for candidate $A$ a cause of $A$
winning, every vote for $B$ is also a cause of $A$ winning.  
To see this, consider a contingency where 8 of the voters for $A$ switch
to $C$.  Then if one of the voters for $B$ votes for $C$, the result is
 a tie;
if that voter switches back to $B$, then $A$ wins (even if some subset
of the voters who switch from $A$ to $C$ switch back to $A$). 

Is this reasonable?  What makes it seem particularly unreasonable is
that if it had just been a contest between $A$ and $B$, with the vote
17--2, then the voters for $B$ would not have been causes of $A$
winning.  Why should adding a third option make a difference?

In some cases it does seem reasonable that adding a third option makes a
difference.  For example, we speak of Nader costing Gore a victory over
Bush in the 2000 election.  But, as Livengood \citeyear{Liv13} points
out, we don't speak of Gore costing Nader a victory, although in a naive
HP model of the situation, all the voters for Gore are causes of Nader
not winning as much as the voters for Nader are causes of Gore not
winning.  The discussion above points a way out of this dilemma.
If a sufficiently large proportion of Bush and Gore voters are taken to
be such strong supporters that they will never change their minds, 
and we make their votes exogenous, then it is still the
case that Nader caused Gore to lose, but not the case that Gore caused 
Nader to lose.  
Similar considerations apply in the case of the 17--2 vote.
(Again, we can use normality considerations to give 
arguably more natural models of these examples.)%
\footnote{As a separate matter, most people would agree that Nader
entering the race was a cause of Gore not winning, while Gore entering
the race was not a cause of Nader not winning.  Here the analysis is
different.   If Nader hadn't entered, it seems reasonable to assume that
there would have been no other strong third-party candidate, so just
about all of Nader's votes would have gone to Bush or Gore, with the
majority going to Gore.  On the other hand, if Gore hadn't entered,
there would have been another Democrat in the race replacing him, and
most of Gore's votes would have gone to the new Democrat in the race,
rather than Nader.}

\commentout{
The last example is Hitchcock's  \citeyear{Hitchcock07} ``bogus
prevention'' example, based on an example due
to Hiddleston \citeyear{Hiddleston05}.  This has been used as the
paradigmatic argument for needing more than just structural equations to
capture causality.  It was already pointed out that bogus prevention
could be handled using the original HP model, without using normality
considerations, by constructing the appropriate model \cite{HH11}.  I
now strengthen this by observing the standard model arguing for the need
for normality is inappropriate.

\xam\label{xam:bogus}
Assassin is in possession of a lethal poison, but has a last-minute
change of heart and refrains from putting it in Victim's coffee.
Bodyguard puts antidote in the coffee, which would have neutralized the
poison had there been any.  Victim drinks the coffee and survives.  
Is Bodyguard's putting in the antidote a cause of Victim surviving?
Most people would say no, but according to the 
HP definition, it is \emph{provided that we model the problem in the obvious
way}, using variables $A$ (which is 1 if Assassin poisons the coffee; 0
otherwise), $B$ (which is 1 if Bodyguard puts in the antidote; 0
otherwise), and $V$ (which is 1 if victim survives; 0 otherwise), and
setting $V = (A \land B) \lor \neg A$.

The similarity of this example to the previous one should already make
clear what is coming next.  We can consider an alternate story where 
the victim survives only if (a) Assassin puts in the poison and Bodyguard
puts in the antidote, (b) Assassin doesn't put in the poison and
Bodyguard puts in the antidote; or (c) Assassin doesn't put in the poison
and Bodyguard puts in the antidote.  The argument now proceeds much as in
the previous example (including constructing the two models that
distinguish the stories).
\exam
}

\section{Do we need AC2(b)?}\label{sec:Hopkins}

In this section, I consider the extent to which we can use AC2(b$'$)
rather than AC2(b), and whether this is a good thing.  

\subsection{The Hopkins-Pearl example}
I start by examining the Hopkins-Pearl example that was intended to show
that AC2(b$'$) was inappropriate.
The following description is taken from \cite{HP01b}.

\xam\label{xam:HopkinsP} Suppose that a prisoner dies 
either if $A$ loads $B$'s gun and $B$ shoots, or if $C$ loads and shoots
his gun.  Taking $D$ to represent the prisoner's death and making the
obvious assumptions about the meaning of the variables, we have that
$D= (A\land B) \lor C$.  Suppose that in the actual
context $u$, $A$ loads $B$'s gun, $B$ does not shoot, but $C$ does load
and shoot his gun, so that the prisoner dies.  That is, $A=1$, $B=0$,
and $C=1$. Clearly $C=1$ is a cause of $D=1$. 
We would not want to say that $A=1$ is a cause of $D=1$,
given that $B$ did not shoot (i.e., given that $B=0$).  However,
suppose that we take the obvious model 
with the random variables
$A$, $B$, $C$, $D$.  
With AC2(b$'$), $A=1$ is a cause of $D=1$.  For we can take
$\vec{W} = \{B,C\}$ and consider the contingency where $B=1$ and $C=0$.
It is easy to check that AC2(a) and AC2(b$'$) hold for this contingency,
so under the original HP definition, $A=1$ is a cause of $D=1$.  However,
AC2(b) fails in this case, since $(M,u) \sat [A \gets 1, C \gets 0](D=0)$.
The key point is that AC2(b) says that for $A=1$ to be a cause of $D=1$, it must
be the case that $D=0$ if only some of the values in $\vec{W}$ are set
to $\vec{w}$.  That means that the other variables 
get the same value as they do in the actual context; in this case, by
setting only $A$ to 1 and leaving $B$ unset, $B$ takes on its original
value of 0, in which case $D=0$.  AC2(b$'$) does not consider this case.

\commentout{
But now consider another story with the same equations: but in this new
story, $D=1$ exactly if either (a) $A=1 \land B=0 \land C=1$, 
(b) $A=0 \land B=1 \land C=1$, (c) $A=0 \land B=0 \land C=1$, or (d)
$A=1 \land B=1$.  Again, the structural equations describing $D$ in
terms of $A$, $B$, and $C$ are the same as in the original story.  
Nevertheless, this second story is intended to be understood differently
from the first.  It should be viewed as saying that death can result
only if (a) $A$ loads and $B$ and $C$  both shoot; (b) $A$ doesn't load and $B$
and $C$ both shoot; (c) $A$ doesn't load, $B$ doesn't shoot, and $C$
shoots;  or (d) $A$ loads and $B$ shoots.  Of course, this story is
inconsistent with our understanding of the way the world works, but if
all we have are the equations, then we cannot rule out the world working
this way.  In any case, with the second story, it seems more reasonable
to say that $B=0$ is a cause of $D=1$.  The only reason that $D$ dies is
because clause (c) applies; if $B$ doesn't shoot, then clause (c) does
not apply.  

Again, we can construct causal models that disambiguate the two
stories.   For the original Hopkins-Pearl story, add a new variable $E$}

Nevertheless, as pointed out by Halpern and Hitchcock \citeyear{HH11}, we can use
AC2(b$'$) if we have the ``right'' model.
Suppose that we add a new variable $E$
such that $E = A \land B$, so that $E=1$ iff $A=B=1$, and set $D =
E \lor C$.  Thus, we have captured the intuition that there are two ways
that the prisoner dies.  Either $C$ shoots, or $A$ loads and $B$ fires
(which is captured by $E$).
It is easy to see that (using either AC2(b) or AC2(b$'$))
$B=0$ is \emph{not} a cause of $D=1$.
\exam

As I now show, the ideas of this example generalize.  
But before doing that, I define the notion of a conservative extension.

\subsection{Conservative extensions}

In the rock-throwing example, 
adding the extra variables converted $\BT = 1$ from being a cause to not
being a cause of $\BS=1$.  Similarly, adding extra variables affected
causality in all the other examples above.  
Of course, without any constraints, it is easy to add variables to get
any desired result.  
For example, consider the rock-throwing model $M_{\RT}'$.
Suppose that we add a variable $\BH_1$ with
equations that set 
$\BH_1 = \BT$ and $\BS = \SH \lor \BT \lor \BH_1$.  This results in a
new ``causal path'' from $\BT$ to $\BS$ going through $\BH_1$,
independent of all other paths.  Not surprisingly, in this model,
$\BT=1$ is indeed a cause of $\BS=1$.

But this seems like cheating.  Adding this new causal path
fundamentally changes the scenario; Billy's throw has a new way of
affecting whether or not the bottle shatters.  While it seems
reasonable to refine a model by adding new information, we want to
do so in a way that does not affect what we know about the old variables.
Intuitively, suppose
that we had a better magnifying glass and could look more carefully at
the model.  We might discover new variables that were previously
hidden.  But we want it to be the case that any setting of the old
variables results in the same observations.    That is, while adding the
new variable refines the model, it does not fundamentally change it.
This is made precise in the following definition.

\dfn\label{dfn:conservative} A causal model $M' = ((\U',\V',\R'),\F')$
is a \emph{conservative 
extension} of 
$M = ((\U,\V,\R),\F)$ if $\U = \U'$, $\V \subseteq \V'$, and, for
all contexts $\vec{u}$,  all variables $X \in \V$, and
all settings $\vec{w}$ of the 
variables in $\vec{W} = \V - \{X\}$, 
we have $(M, \vec{u}) \sat [\vec{W} \gets \vec{w}](X=x)$ iff
$(M', \vec{u}) \sat [\vec{W} \gets \vec{w}](X=x)$.
That is, no matter how we set the variables other than $X$, $X$ has the 
same value in context $\vec{u}$ in both $M$ and $M'$.
\edfn

According to the definition, $M'$ is a
conservative extension of $M$ iff, for certain formulas $\psi$ involving
only variables in $\V$, 
namely, those of the form $[\vec{W} \gets \vec{w}](X=x)$,
$(M,u) \sat \psi$ iff $(M',u) \sat \psi$.  As 
the following lemma shows, this is actually true for all formulas
involving only variables in $\V$, not just ones of a special form.

\lem\label{lem:conservative}
 Suppose that $M'$ is a conservative extension of $M = ((\U,\V,\R),\F)$.
Then for all causal formulas $\phi$ that mention only variables in $\V$
and all contexts $\vec{u}$, we have 
$(M,\vec{u}) \sat \phi$ iff $(M',\vec{u}) \sat \phi$.
\elem

\prf \shortv{See the full paper.  \eprf}
\fullv{Since $M$ is a recursive model, there is some partial
order $\preceq$ on the endogenous variables
such that unless $X \preceq Y$, 
$Y$ is independent of $X$ in $M$; that is, unless $X
\preceq Y$, changing the value of $X$
has no impact on the value of $Y$ according to the structural
equations in $M$, no matter what the setting of the other
variables.  It is almost immediate from the definition of conservative
extension that, 
for all $X, Y \in \V$,  $Y$ is independent of $X$ in $M$
iff $Y$ is independent of $X$ in $M'$.  Also note that if 
$X \preceq Y$, then it is not the case that $Y
\preceq X$, so if $X \preceq Y$, then $X$ is
independent of $Y$ (in both $(M,\vec{u})$ and $(M',\vec{u})$).  
Say that $X$ is independent of a set $\vec{W}$ of endogenous variables
in $(M,\vec{u})$ if $X$ is independent of $Y$ in $(M,\vec{u})$ for all
$Y \in \vec{W}$. 

Suppose that $\V = \{X_1, \ldots, X_n\}$.  
Since $M$ is a recursive model, we can assume without loss of generality
that these variables are ordered so that $X_1 \prec^M \cdots \prec^M X_n$.
I now prove by induction on $j$ that, for all $\vec{W} \subseteq \V$,
all settings $\vec{w}$ of the variables in $\vec{W}$, all contexts $\vec{u}$,
and all $x_j \in 
\R(X_j)$, we have $(M,\vec{u}) \sat [\vec{W} \gets \vec{w}](X_j = x_j)$
iff $(M',\vec{u}) \sat [\vec{W} \gets \vec{w}]( X_j = x_j)$.

For the base case of the induction, given $\vec{W}$, let $\vec{W}' = \V
- (\vec{W} \union \{X_1\})$, and let $\vec{w}'$ be an arbitrary setting
of the variables in $\vec{W}'$.  Then  we have
$$\begin{array}{lll}
&(M,\vec{u}) \sat [\vec{W} \gets \vec{w}] (X_1= x_1)\\
\mbox{iff } &(M,\vec{u}) \sat [\vec{W} \gets \vec{w}, \vec{W}'\gets
\vec{w}'] (X_1 = x_1) &\mbox{[since $X$ is independent of $\vec{W}$ in
    $(M,\vec{u})$]}\\
\mbox{iff } &(M',\vec{u}) \sat [\vec{W} \gets \vec{w}, \vec{W}'\gets
\vec{w}'] (X_1 = x_1) &\mbox{[since $M'$ is a conservative extension of
$M$]}\\
\mbox{iff } &(M',\vec{u}) \sat [\vec{W} \gets \vec{w}]
(X_1 = x_1) &\mbox{[since $X$ is independent of $\vec{W}$ in $(M',\vec{u})$].}
\end{array}
$$

This completes the proof of the base case.  Suppose  that $1< j < n$
and the result holds for $1,\ldots, 
j-1$; I prove it for $j$.  
Given $\vec{W}$, now let $\vec{W}' = \V
- (\vec{W} \union \{X_{j}\})$, let $\vec{W}'_1 = \vec{W}' \inter
\{X_1, \ldots, X_{j-1}\}$, and let $\vec{W}'_2 = \vec{W}' - \vec{W}_1'$.
Choose $\vec{w}'_1$ such that $(M,\vec{u}) \sat 
[\vec{W} \gets \vec{w}](\vec{W}'_1 = \vec{w}'_1)$.  Since 
$\vec{W}_1' \subseteq \{X_1, \ldots, X_{j-1}\}$, by the induction
hypothesis, $(M',\vec{u}) \sat 
[\vec{W} \gets \vec{w}](\vec{W}'_1 = \vec{w}'_1)$.  
It easily follows that  we have 
$(M,\vec{u}) \sat [\vec{W} \gets \vec{w}] (X_{j} = x_{j})$ iff
$(M,\vec{u}) \sat [\vec{W} \gets \vec{w}, \vec{W}_1' \gets \vec{w}_1']
(X_{j} = x_{j} )$, and similarly for $M'$.
Thus,
$$\begin{array}{lll}
&(M,\vec{u}) \sat [\vec{W} \gets \vec{w}] (X_{j} = x_{j})\\
\mbox{iff } &(M,\vec{u}) \sat [\vec{W} \gets \vec{w},\vec{W}_1' \gets
\vec{w}_1'] (X_{j} = x_{j}) &\mbox{[as observed above]}\\ 
\mbox{iff } &(M,\vec{u}) \sat [\vec{W} \gets \vec{w}, \vec{W}_1'\gets
\vec{w}_1',\vec{W}_2' \gets \vec{w}_2'] (X_{j} = x_{j}) &\mbox{[since $X_j$
in independent of $\vec{W}_2'$ in $(M,\vec{u})$]}\\
\mbox{iff } &(M',\vec{u}) \sat [\vec{W} \gets \vec{w}, \vec{W}_1'\gets
\vec{w}_1', \vec{W}_2' \gets \vec{w}_2'] (X_{j} = x_{j}) &\mbox{[since $M'$
is a conservative extension of $M$]}\\
\mbox{iff } &(M',\vec{u}) \sat [\vec{W} \gets \vec{w}, \vec{W}_1'\gets
\vec{w}_`'] (X_{j} = x_{j}) &\mbox{[since $X_j$ is independent of 
$\vec{W}_1'$ in $(M',\vec{u})$]}\\
\mbox{iff } &(M',\vec{u}) \sat [\vec{W} \gets \vec{w}] (X_{j} =
x_{j})
&\mbox{[as observed above].}
\end{array}
$$
This completes the proof of the inductive step.

Since, in general $(M,u) \sat [\vec{W} \gets \vec{w}] (\psi_1 \land
\psi_2)$ iff $(M,u) \sat [\vec{W} \gets \vec{w}] \psi_1 \land
[\vec{W} \gets \vec{w}] \psi_2$ and 
$(M,u) \sat [\vec{W} \gets \vec{w}] \neg \psi_1$ iff 
$(M,u) \sat \neg [\vec{W} \gets \vec{w}] \psi_1$, and similarly for
$M'$, an easy induction shows that 
$(M,u) \sat [\vec{W} \gets \vec{w}] \psi$ iff 
$(M',u) \sat [\vec{W} \gets \vec{w}] \psi$ for an arbitrary Boolean
combination $\psi$ of primitive events that mentions only variables in
$\V$.  Another easy induction shows
that 
$(M,u) \sat \psi$ iff
$(M',u) \sat \psi$ for all causal formulas $\psi$.
\eprf
}

\subsection{Avoiding AC2(b)}

I now show that we can always use AC2(b$'$) instead of AC2(b), if we add
extra variables.

\thm\label{thm:AC2b}
 If $X=x$ is  not a cause of $Y=y$ in $(M,\vec{u})$ using
AC2(b), but is a cause using AC2(b$'$),
then there is a
model $M'$ that is a conservative extension of $M$ such that $X=x$ is
not a cause of $Y=y$ using AC2(b$'$). 
\ethm


\prf Suppose that 
$(\vec{W},\vec{w},x')$ is a witness to $X=x$ being a cause of $Y=y$ in
$(M,\vec{u})$ using AC2(b$'$).  
Let  $(M,\vec{u}) \sat \vec{W} 
= \vec{w}^*$.  We must have $\vec{w} \ne \vec{w}^*$, 
for otherwise it is easy to see that $X=x$ would be a cause of $Y=y$ in
$(M,\vec{u})$ using AC(2b) with witness $(\vec{W},\vec{w},x')$.

If $M'$ is a conservative extension of $M$ with additional variables
$\V'$, 
say that
$(\vec{W'}, \vec{w}', x')$ \emph{extends} 
$(\vec{W},\vec{w},x')$ if $\vec{W} \subseteq \vec{W}' \subseteq \vec{W}
\union \V'$ and $\vec{w}'$ agrees with $\vec{w}$ on the variables in
$\vec{W}$.  

I now construct a conservative extension $M'$ of $M$ in which $X=x$ is
not a cause of $Y=y$ using AC2(b$'$) with a witness extending
$(\vec{W'},\vec{w},x')$. 
Of course, this just kills one witness.  I then show that we can
construct further extensions to kill all other witnesses to $X=x$ being
a cause of $Y=y$ using AC2(b$'$).   

Let $M'$ be obtained from $M$ by adding one new variable
$\NW$.  All the variables have the same equations in $M$
and $M'$ except for $Y$ and (of course) $\NW$.  
The equations for $\NW$ are easy to explain:
if $X=x$ and $\vec{W} = \vec{w}$, then $\NW = 1$;
otherwise, $\NW = 0$.  The equations for $Y$ are the
same in $M$ and $M'$ (and do not depend on the value of
$\NW$) except for two special cases.  To define these
cases, for each variable $Z \in \V - \vec{W}$, if $x'' \in \{x,x'\}$,
define $z_{x'',\vec{w}}$ as the 
value such that $(M,\vec{u}) \sat [X \gets x'', \vec{W} \gets \vec{w}](Z =
z_{x'',\vec{w}})$.  That is, $z_{x'',\vec{w}'}$ is the value taken by $Z$ if
$X$ is set to $x''$ and $\vec{W}$ is set to $\vec{w}$. 
Let $\vec{V}'$ consist of all variables in $\V$ other than $Y$,
let $\vec{v}'$ be a setting of the variables in $\vec{V}'$, and let 
$\vec{Z}'$ consist of all variables in $\vec{V}' - \vec{W}$ other than $X$.
Then we want the
equations for $Y$ in $M'$ to be such that for all $j \in \{0,1\}$, we
have 
$$\begin{array}{ll}
(M,\vec{u})\sat [\vec{V}' \gets \vec{v}'](Y=y'') \mbox{ iff } \\
(M',\vec{u}) \sat [\vec{V}' \gets \vec{v}'; \NW \gets
j](Y=y'')
\end{array}$$
unless the assignment $\vec{V}' \gets \vec{v}'$
results in either  (a) $X=x$, $\vec{W} = \vec{w}$, $Z = z_{x,\vec{w}}$
for all $Z \in \vec{Z}'$, and
$\NW = 0$ or (b) $X=x'$, $\vec{W} = \vec{w}$, 
$Z = z_{x',\vec{w}}$ for all $Z \in \vec{Z}'$, and
$\NW = 1$.  (Note that in both of these cases, the value
of $\NW$ is ``abnormal''.  If $X=x$, $\vec{W} =
\vec{w}$ and $Z = z_{x,\vec{w}}$ for all $z \in \vec{Z}'$, then
$\NW$ should be 1; if we set
$X$ to $x'$ and change the values of the variables in $\vec{Z}'$
accordingly, then $\NW$ should be 0.)
If (a) holds, $Y= y'$ in $M'$; if (b) holds,
$Y = y$.

\fullv{I now show that $M'$ has the desired properties and, in addition, does
not make $X=x$ a cause in new ways.}

\lem\label{lem:AC2}
\begin{itemize}
\item[(a)] It is not the case that $X=x$ is a cause of $Y=y$
using AC2(b$'$) in $(M',\vec{u})$ with a witness that extends
$(\vec{W},\vec{w},x')$.
\item[(b)] $M'$ is a conservative extension of $M$.
\item[(c)] If $X = x$ is a cause of $Y=y$ in $(M',\vec{u})$
using AC2(b) (resp. AC2(b$'$)) with a witness extending
$(\vec{W}',\vec{w}',x'')$ then 
$X = x$ is a cause of $Y=y$ in $(M,\vec{u})$
using AC2(b) (resp. AC2(b$'$)) with witness 
$(\vec{W}',\vec{w}',x'')$.
\end{itemize}
\elem

\shortv{The proof of Lemma~\ref{lem:AC2} and the remainder of the proof
of Theorem~\ref{thm:AC2b} can be found in the full paper. \eprf}

\fullv{\prf For part (a), 
suppose, by way of contradiction, that $X=x$ is a cause of $Y=y$
using AC2(b$'$) in $(M',\vec{u})$ with a witness $(\vec{W}',\vec{w}',x')$ that
extends $(\vec{W},\vec{w},x')$.  If $\NW \notin \vec{W}'$,  
then $\vec{W}' = \vec{W}$.   But then, since $(M',\vec{u}) \sat \NW =
0$ and 
$(M', \vec{u}) \sat [X \gets x; \vec{W} \gets \vec{w},
\NW = 0](Y=y')$, 
it follows that 
$(M', \vec{u}) \sat [X \gets x; \vec{W} \gets \vec{w}](Y=y')$,  so
AC2(b$'$) fails, contradicting the assumption that
$X=x$ is a cause of $Y=y$.
Now suppose that $\NW  \in \vec{W}'$.  There are two
cases, depending on how the value of $\NW$ is set in
$\vec{w}'$.  
If $\NW = 0$, then again, since $(M',\vec{u}) \sat [X \gets x,
\vec{W} \gets w, \NW \gets 0](Y=y')$, AC2(b$'$) fails;
and 
if $\NW = 1$, then since $(M',\vec{u}) \sat [X \gets x',
\vec{W} \gets \vec{w}, \NW \gets 1](Y=y)$,  AC2(a) fails.
So, in all cases, we get a contradiction to the assumption that 
$X=x$ is a cause of $Y=y$
using AC2(b$'$) in $(M',\vec{u})$ with a witness $(\vec{W}',\vec{w}',x')$ that
extends $(\vec{W},\vec{w},x')$.

For part (b), note that the only
variable in $\V$ for which the equations in $M$ and $M'$ are different
is $Y$. Consider any setting of the
variables in $\V$ 
other than $Y$.  Except for the two special cases noted above, the
value of $Y$ is clearly the same in $M$ and $M'$.  But for these two
special cases, as was noted above, the value of $\NW$ is
``abnormal'', that is, it is 
not the same as its 
value according to the equations given the setting of the other
variables.  It follows that for all 
settings $\vec{v}$ of the variables $\vec{V}'$ in $\V$ other than $Y$ and all
values $y''$ of $Y$,  
we have $(M,\vec{u}) \sat (\vec{V}' \gets \vec{v}](Y = y'')$ iff 
$(M',\vec{u}) \sat (\vec{V}' \gets \vec{v}](Y = y'')$.
Thus, $M'$ is a conservative extension of $M$.

For part (c), suppose that $X = x$ is a cause of $Y=y$ in $(M',\vec{u})$
using AC2(b) (resp. AC2(b$'$)) with witness
$(\vec{W}'',\vec{w}'',x'')$.  Let $\vec{W}'$ and $\vec{w}'$ be the
restrictions of $\vec{W}''$ and $\vec{w}''$, respectively, to the variables
in $\V$.  If $\NW \notin \vec{W}''$ (so that $\vec{W}'' = \vec{W}'$)
then, since $M'$ is a conservative extension of $M$, it easily follows
that $(\vec{W}', \vec{w}',x'')$ is a witness to $X=x$ being a cause of
$Y=y$ in $(M,\vec{u})$ using AC2(b) (resp. AC2(b$'$)).  If $\NW \in
\vec{W}''$, it suffices to show that $(\vec{W}',\vec{w}',x'')$ is also
a witness to $X=x$ being a cause of $Y=y$ in $(M',\vec{u})$; that is,
$\NW$ does not play an essential role in the witness.   I now do this.

If $\NW=0$ is a conjunct of $\vec{W}''=\vec{w}''$,
since the equations for $Y$ are the same in $M$ and
$M'$ except for two cases, the only way that $\NW=0$ can play an
essential role  in the witness
is if setting $\vec{W}' = \vec{w}'$
and $X = x$ results in $\vec{W} = \vec{w}$ and $Z = z_{x,\vec{w}}$ for
all $Z \in \vec{Z}'$ (i.e., we are in the first of
the two cases where the value of $Y$ does not agree in $(M,\vec{u})$ and
$(M',\vec{u})$).   But then $Y=y'$, so if this were the case,
AC2(b) (and hence AC2(b$'$)) would not hold.  Similarly, if $\NW=1$
is a conjunct of 
$\vec{W}'' = \vec{w}''$, $\NW$ plays a role only if $x'' = x'$ and
setting $\vec{W}' = \vec{w}'$ and $X = x'$ results in 
results in $\vec{W} = \vec{w}$ and $Z = z_{x',\vec{w}}$ for
all $Z \in \vec{Z}'$ (i.e., we are in the second of
the two cases where the value of $Y$ does not agree in $(M,\vec{u})$ and
$(M',\vec{u})$).   But then $Y=y$, so if this were the case,
AC2(a) would not hold, and again we would have
a contradiction to $X=x$ being a cause of $Y=y$ in $(M',\vec{u})$ with
witness $(\vec{W}'',\vec{w}'',x'')$.
Thus, $(\vec{W}',\vec{w}',x'')$ must be a witness to $X=x$ being cause
of $Y=y$ in $(M',\vec{u})$, and hence also in $(M,\vec{u})$.
This completes the proof of part (c). 
\eprf

Lemma~\ref{lem:AC2} is not quite enough to 
complete the proof of
Theorem~\ref{thm:AC2b}.  There may be several witnesses to $X=x$ being a
cause of $Y=y$ in $(M,\vec{u})$ using AC2(b$'$).  Although we have removed one
of the witnesses, some others may remain, so that $X=x$ may still be a
cause of $Y=y$ in $(M',\vec{u})$.  But by Lemma~\ref{lem:AC2}(c), if there is a
witness to $X=x$ being a cause of $Y=y$ in $(M',\vec{u})$, it must extend a
witness to $X=x$ being a cause of $Y=y$ in $(M,\vec{u})$.  We can repeat the
construction of Lemma~\ref{lem:AC2} to kill this witness as well. 
Since there are only finitely many witnesses to $X=x$ being a cause of $Y=y$ in
$(M,\vec{u})$, after finitely many extensions, we can kill them all.  After this
is done, we have a causal model $M^*$ extending $M$ such that $X=x$ is
not a cause of $Y=y$ in $(M^*,\vec{u})$ using AC2(b$'$).
\eprf

It is interesting to apply the construction of Theorem~\ref{thm:AC2b} to 
Example~\ref{xam:HopkinsP}.  The variable $N$ added by
the construction is almost identical to $E$.  Indeed, the only
difference is that $\NW=0$  if $A=B=C=1$, while $E=1$ in this case.
But since $D=1$ if $A=B=C=1$ and $\NW=0$, the equations for $D$ are the
same in both causal models if $A=B=C=1$.  While it seems strange, given
our understanding of the meaning of the variables, to have $N=0$ if
$A=B=C=1$, it is easy to see that this definition works equally well in
showing that $A=1$ is not a cause of $D=1$ using AC2(b$'$) in the
context where $A=1$, $B=0$, and $C=1$. 
}

\subsection{Discussion}
Theorem~\ref{thm:AC2b} suggests that, by adding extra variables
appropriately, we can go back to 
the definition of causality using AC2(b$'$) rather than AC2(b).  This
has some technical advantages.  For example, with AC2(b$'$), causes are
always single conjuncts \cite{EL01,Hopkins01}.  As shown in \cite{Hal39},
this is not in general the case with AC2(b); it may be that $X_1 = x_1
\land X_2 = x_2$ is a cause of $Y=y$ with neither $X_1=x_1$ nor $X_2 =
x_2$ being causes (see also Example~\ref{xam:non-causal}).  
It also seems that testing for causality is harder
using AC2(b). Eiter and Lukasiewicz \citeyear{EL01} show that, using 
AC2(b$'$), testing for causality is NP-complete for binary models
(where all random variables are binary) and $\Sigma_2$-complete in
general; with AC2(b), it seems to be $\Sigma_2$-complete in the binary
case and  $\Pi_3$-complete in the general case \cite{ACHI14}.

On the other hand, adding extra variables may not always be a natural
thing to do.  For example, in Beer et al.'s 
\citeyear{BBCOT12} analysis of software errors using causality,   
the variables chosen for the analysis are determined by
the program specification.  Moreover, Beer et al.~give examples where
AC2(b) is needed to get the intuitively correct answer.   Unless we are
given a principled way of adding extra variables so as to be able to
always use AC2(b$'$), it is not clear how to automate an analysis.
\fullv{
In addition, as we saw above, adding the 
extra variable $N$ as in Theorem~\ref{thm:AC2b} rather than $E$ 
result in an ``unnatural'' model.  }
There does not always seem to be
a ``natural'' way of adding extra variables
so that AC2(b$'$) suffices (even assuming that we can agree on 
what
``natural'' means!).  

Adding extra variables also has an impact on complexity.  Note that, in
the worst case, we may have to add an extra variable for each pair
$(W,w)$ such that there is a witness $(W,w,x')$ for $X=x$ being a cause
of $Y=y$.  In all the standard examples, there are very few witnesses
(typically 1--2), but I have been unable to prove a nontrivial bound on
the number of witnesses.  

More experience is needed to determine which of AC2(b) and AC2(b$'$) is 
most appropriate.  Fortunately, in many cases, the causality judgment is
independent of which we use.

\fullv{
\section{Normality}\label{sec:normality}

As was already observed in \cite{HH11}, the example that
motivated the use of normality considerations can also be dealt with by 
adding variables to the model in an arguably reasonable way.  Consider
the following example, 
given by Hitchcock \citeyear{Hitchcock07}, based on an example due
to Hiddleston \citeyear{Hiddleston05}.

\xam\label{xam:bogus}
Assassin is in possession of a lethal poison, but has a last-minute
change of heart and refrains from putting it in Victim's coffee.
Bodyguard puts antidote in the coffee, which would have neutralized the
poison had there been any.  Victim drinks the coffee and survives.  
Is Bodyguard's putting in the antidote a cause of Victim surviving?
Most people would say no, but
according to the HP definition (with either AC2(b) or AC2(b$'$)), it is.
For in the contingency 
where Assassin puts in the poison, Victim survives iff Bodyguard puts
in the antidote.  
\exam

What makes this particularly troubling is that the obvious naive model
is isomorphic to the naive model in the rock-throwing example
(illustrated in Figure~\ref{fig:rt}).  Specifically,
if we take 
$A$ (for ``assassin does not put in poison''), $B$ (for ``bodyguard puts
in antidote''), and $\VS$ (for ``victim survives''),
then $\VS = A \lor B$, just as $\BS = \ST \lor \BT$.  However, while most
people agree that $\ST = 1$ a cause of $\BS = 1$ in this, they do 
not view $A=1$ as a cause of $\VS = 1$.
Using normality considerations, we can say that $A=1$ is not a cause
because the witness world, where $A=0$, is less normal than the actual
world.  It is not normal to put poison in coffee.  But would we feel
differently in a universe where poisoning occurred frequently, or was
normal in the sense that it was accepted practice?  

Arguably a better solution to this problem, already suggested in
\cite{HH11}, is to add an additional variable.
Suppose we add a variable 
$\PN$ to the model, representing whether a chemical reaction
takes place in which poison is neutralized, where 
$\PN = \neg A \land B$ ($A$ puts in the poison and $B$ puts in the
antidote) and $\VS = A \lor  \PN$, it is easy to
check that now $B=1$ is no longer a cause of $\VS=1$.  Intuitively, the
antidote is a cause of the victim living only if it actually neutralized
the poison.  

Blanchard and Schaffer \citeyear{BS13} have used this example and others
to argue that we do not need to use normality at all in determining
causality.  I do not agree.  As we have seen,
thinking in terms of
normality helps in the Livengood voting example; there are many other
examples given in \cite{HH11} where the use of normality, and in
particular the ability to use normality to allow for gradation of
causality, seems to be helpful.  Moreover, as I mentioned earlier,
people seem to take normality considerations into account.
Finally, in the case of normality, we do not yet have an analogue to
Theorem~\ref{thm:AC2b} 
that says that we can always add extra variables to remove the need for
normality.  There may well be examples where normality solves the
problem, while no number of extra variables will deal with it.
}

\section{The Stability of (Non-)Causality}\label{sec:stability}
The examples in Section~\ref{sec:examples}
raise a potential concern.  Consider the rock-throwing example again.
Adding extra variables changed $\BT=1$ from being a cause of $\BS=1$ to
not being a cause.  Could
adding even more variables convert $\BT = 1$ back to being a cause?
Could it then alternate further?  

These questions of stability have been raised before.
Strevens \citeyear{Strevens08} provides an example where what Strevens
calls a cause can become a non-cause if extra
variables are added according to Woodward's \citeyear{Woodward03}
definition of causality;%
\footnote{Actually, Strevens considered what Woodward called a
  \emph{contributing cause}.}  Eberhardt \citeyear{Eberhardt14} shows that
this can also happen for \emph{type causality} (``smoking causes
cancer'' rather than ``Mr T.'s smoking for 20 years caused him to get
cancer'') using Woodward's definition.  Here I consider the situation
in more detail for the HP definition and show that it can get much worse.
In general, we can convert an event from being a
cause to a non-cause and then back again infinitely often.  

Consider an arbitrary model $M$ with variables $A$ and $B$ and a context $u$
such that $(M,u) \sat 
A=1 \land B=1$, but $A=1$ is not a cause of $B=1$ in $(M,u)$.  I now show how to
extend $M$ in a conservative way so as to make $A=1$ a cause of $B=1$.
Add a new binary variable to $M$, say $X_1$, to get a model $M'$.
Normally $X_1 = 1$. 
The equations for all variables are the 
same in $M$ and $M'$ unless $A=X_1 = 0$.
If $A=X_1=0$, then $B=0$.  But if
$A=1$ then $B=1$, no matter what the value of $X_1$.  It easily
follows that $A=1$ is a cause of $B=1$, with witness $(\{X_1\},0,0)$.
It is then not hard to then add a variable $Y_1$ to ``neutralize'' the
effect of $X_1$, so that $A=1$ is not a cause of $B=1$.   Repeating this
construction infinitely often, we get a sequence of models where the
the answer to the question of whether $A=1$ is a cause of $B=1$ alternates
infinitely often.

I now formalize this.  Specifically, I construct a
sequence $M_0, M_1, M_2, \ldots$ of causal models and 
a context $u$ such that
$M_{n+1}$ is a conservative extension $M_n$, $A=1$ is not a
cause of $B=1$ in the causal settings $(M_n,u)$ where $n$ is even
and $A=1$ is a cause of $B=1$ in
the the causal settings $(M_n,u)$ where $n$ is odd.
That is, the answer to the question ``Is $A=1$ a cause of $B=1$?''
alternates as we go along the sequence of models.  

$M_0$ is just the model with two binary endogenous variables $A$ and $B$ with
one binary exogenous variable $U$.  The variables $A$ and $B$ are
independent of each other; their value is completely determined by the
context.  In the context $u_1$ where $U=1$, $A=B=1$.  In the context
$u_0$ where $U=0$, $A=B=0$.  Clearly, $A=1$ is not a cause
of $B=1$ in $(M_0,u_1)$.  

The models $M_1, M_2, M_3, \ldots$ are defined inductively.
For $n \ge 0$, we get 
$M_{2n+1}$ from $M_{2n}$ by adding a new variable $X_{n+1}$; we get 
$M_{2n+2}$ from $M_{2n+1}$ by adding a new variable $Y_{n+1}$.  Thus, for $n
\ge 0$, the model $M_{2n+1}$ has the endogenous variables $A, B, X_1,
\ldots, X_{n+1}, Y_1, \ldots, Y_{n}$ and  the model $M_{2n+2}$ has the 
endogenous variables $A, B, X_1, \ldots, X_{n+1}, Y_1, \ldots, Y_{n+1}$. 
All these models have just one binary exogenous variable $U$.  
For $n \ge 0$, the
exogenous variable determines the value of $A, X_1, \ldots,
X_{n+1}$ in models $M_{2n+1}$ and $M_{2n+2}$; in the context $u_j$,
these variables 
all have value $j$.  
In addition, in $u_0$, $B = 0$, no matter how the other
variables are set. 
If $n \ge 1$, then
in $M_{2n}$ and $M_{2n+1}$, the equation for $Y_j$ is just $Y_j =
X_j$, $j=1, \ldots, n$; $X_j$ determines $Y_j$.  
In $(M_{2n-1},u_1)$,
$B=1$ unless either (a) $A=0$ and either $X_n = 0$ or for some $j <
n$, $X_j = Y_j = 0$, 
or (b) $A=1$ and $X_j \ne Y_j$ for some $j < n$.  
In $M_{2n}$, if $U=1$, then $B=1$ unless
either (a) $A=0$ and for some $j \le n$,
$X_j = Y_j = 0$ or (b) $A=1$ and $X_j \ne Y_j$ for some $j \le n$.  
Intuitively, $B=1$ unless $A=0$ and $X_j$ and $Y_j$ both take on the
exceptional value 0 (or just $X_n$ does, if there is no corresponding $Y_n$),
or $A=1$ and $X_j$ is different from $Y_j$ (which
is also an exceptional circumstance).  

\thm\label{pro:stability}
For all $n \ge 0$, $M_{n+1}$ is a conservative extension of
$M_n$.  Moreover, $A=1$ is not a cause of $B=1$ in $(M_{2n},u_1)$ and
$A=1$ is a cause of $B=1$ in $(M_{2n+1},u_1)$.  
\ethm

\prf
Fix $n \ge 0$.
To see that $M_{2n+1}$ is a conservative extension of $M_{2n}$, note
that for the variables $A, B, X_1, \ldots, X_n, Y_1, \ldots, Y_n$ that
appear in both $M_{2n}$ and $M_{2n+1}$, 
the equations for all variables but $B$ are the same in
$M_{2n}$ and $M_{2n+1}$.  It thus
clearly suffices to show that, no matter what the value of $U$, for
every setting of 
the variables $A$, $X_1, \ldots, X_{n}, Y_1, \ldots, Y_{n}$, the value of 
$B$ is the same in both $M_{2n}$ and $M_{2n+1}$.%
\footnote{Of course, if $n=0$, there are no variables $X_1, \ldots,
  X_n, Y_1,\ldots, Y_n$, so it suffices to show that for all settings
  of $A$, the value of $B$ is the same in $M_0$ and $M_1$.  A similar
  comment applies elsewhere when $n=0$.}
  If $U=0$, $B=0$ in
both $M_{2n}$ and $M_{2n+1}$.  If  $U=1$, in $M_{2n+1}$, no matter how $A,
X_1, \ldots, X_{n}, Y_1, \ldots, Y_{n}$ are set, $X_{n+1} = 1$.  
And if $X_{n+1} = 1$, then the value of $B$ depends on the values of 
$A, X_1, \ldots, X_{n}, Y_1, \ldots, Y_{n}$ in 
$M_{2n+1}$ in the same way that it does in $M_{2n}$. 

The argument that $M_{2n+2}$ is a conservative extension of $M_{2n+1}$
is almost identical.  Now we have to show that, no matter what the
value of $U$,
for every setting of 
the variables $A$, $X_1, \ldots, X_{n+1}, Y_1, \ldots, Y_{n}$, the value of 
$B$ is the same in both $M_{2n+1}$ and $M_{2n+2}$.   Again, this is
immediate if $U=0$.  If $U=1$, since $Y_{n+1} = X_{n+1}$ in $M_{2n+2}$, 
the result again follows easily.  

To see that $A=1$ is a cause of $B=1$ in $(M_{2n+1},u_1)$, take $\vec{W}
=\{X_{n+1}\}$.  It is immediate that $(M_{2n+1},u_1) \sat [A \gets 0, X_{n+1} \gets
  0](B=0)$, so AC2(a) holds.  Moreover, $(M_{2n+1},u_1) \sat [A \gets 1,
  X_{n+1} \gets 
  0](B=1)$ and $(M_{2n+1},u_1) \sat [A \gets 1, X_{n+1} \gets 1](B=1)$, so AC2(b)
holds.  (Note that all of $X_1, \ldots, X_{n}, Y_1, \ldots, Y_{n}$
get the same value if the context  $U=1$ whether or not $X_{n+1}$ is set to 1.)  

Finally, to see that $A=1$ is not a cause of $B=1$ in
$(M_{2n},u_1)$, suppose, by way of contradiction, it is a cause, with
witness $(\vec{W}, \vec{w},0)$.  For AC2(a) to hold, we must have
$(M_{2n},u_1) \sat [A\gets 0,\vec{W}\gets \vec{w}](B=0)$.  Thus, 
there must some $j < n$ such that $\{X_j,Y_j\} \subseteq
  \vec{W}$ and $\vec{w}$ is such that $X_j$ and $Y_j$ are set to 0.
  But then let $\vec{W}' = \vec{W} - \{X_j\}$.  Then we must have 
$(M_{2n},u_1) \sat [A\gets 1,\vec{W}'\gets \vec{w}](B=0)$, because
  if $U=1$, then $X_j$ is set to 1, and this is not overridden by
  $\vec{W}'$, and if $A=1$, $X_j=1$, and $Y_j = 0$, then $B=0$.  Thus,
  AC2(b) does not hold.  This completes the argument.
\eprf

Theorem~\ref{pro:stability} is somewhat disconcerting.  
It seems that looking more and more carefully at a situation should not
result in our view of $X=x$ being a cause of $Y=y$ alternating
between ``yes'' and ``no'', at least, not if we do not discover anything
inconsistent with our understanding of the relations between previously
known variables.  
Yet, Theorem~\ref{pro:stability} shows that this can happen.
Moreover, the construction used in Theorem~\ref{pro:stability} can be
applied to \emph{any} model $M$ such that $(M,\vec{u}) \sat A=1 \land
B=1$, but $A$ and $B$ are independent of each other (so that, in
particular, $A=1$ is not a cause of $B=1$), to get a sequence of models
$M_0, M_1, \ldots$, with $M= M_0$ and $M_{n+1}$ a conservative
extension of $M_n$ such that the truth of 
the statement  ``$A=1$ is a cause of $B=1$ in $(M_n,\vec{u})$''
alternates as we go along the sequence.

While disconcerting, I do not believe that, in fact, this is a problem. 
A child may start with a primitive understanding of how the world
works, and believe that just throwing a rock causes a bottle to
shatter.  Later he may  become aware of the importance of the rock
actually hitting the bottle. Still later, he may become of other
features critical to bottles shattering.  This increased awareness can
and should result in causality ascriptions changing.    However, in
practice, there are very few new features that should matter.  We can
make this precise by observing that, most new features that we
become aware of are almost surely irrelevant to the bottle shattering
except perhaps in highly abnormal circumstances.  If the new variables
were relevant, we probably would have become aware of them sooner.  
(Recall the gloss that I gave above when introducing the variable
$X_n$: the value $X_n = 0$, which was needed to establish $A=1$ being
a cause of $B=1$, was an abnormal value.)  

As I now show, once we take normality into account, under reasonable
assumptions, non-causality is stable.
To make this precise, I must first extend the notion of conservative
extension to extended causal models so as to take the normality
ordering into account. 

\dfn\label{dfn:conservative1} An extended causal model $M' =
(\S',\F',\succeq')$ 
is a \emph{conservative 
extension} of an extended causal model
$M = (\S,\F,\succeq)$ 
if the causal model $(\S',\F')$ underlying $M'$ is a conservative
extension of the causal model $(\S,\F)$ underlying $M$ according to
Definition~\ref{dfn:conservative} and, in addition, 
the following condition holds, where $\V$ is the set of endogenous 
variables in $M$:
\begin{itemize}
\item[CE.] For all contexts $\vec{u}$, if $\vec{W} \subseteq
\V$, then $s_{\vec{W} = \vec{w},\vec{u}} \succeq s_{\vec{u}}$
iff $s_{\vec{W} = \vec{w},\vec{u}} \succeq'  s_{\vec{u}}$.
\end{itemize}
\edfn
Roughly speaking, CE say that the normality ordering when restricted
to worlds characterized by settings of the variables in $\V$ is the
same in $M$ and $M'$.  (Actually, CE says less than this. I could
have taken a stronger version of CE that would be closer to this
English gloss: if $\vec{W} \union \vec{W}' \subseteq \V$, then 
$s_{\vec{W} = \vec{w},\vec{u}} \succeq s_{\vec{W}' =   \vec{w}',\vec{u}}$ iff
$s_{\vec{W} = \vec{w},\vec{u}} \succeq' s_{\vec{W}' =
  \vec{w}',\vec{u}}$.  The version of CE that I consider suffices to
prove the results below, but this stronger version seems
reasonable as well.)

For the remainder of this section, I work with extended causal models $M$ and
$M'$, and so 
use the extended HP definition of causality that takes normality into
account, although, for ease of exposition, I do not mention this explicitly.
As above, I take $\succeq$ and $\succeq'$ to be the preorders in $M$
and $M'$, respectively.

I now provide a condition that almost ensures that
non-causality is stable.
Roughly speaking, I want it to be abnormal for 
a variable to take on a value other than that specified by the
equations.
Formally, say that \emph{in world $s$, $V$ takes on a value other than that
  specified by the equations in $(M,\vec{u})$} if,
taking $\vec{W}^*$ to consist of all endogenous variables in $M$ other
than $V$, if $\vec{w}^*$ gives the values of the variables in $\vec{W}^*$ in
$s$, and $v$ is the value of $V$ is $s$, then 
$(M,\vec{u}) \sat [\vec{W}^* \gets \vec{w}^*](V \ne v)$.  For future
reference, note that 
it is easy to check that if $\vec{W} \subseteq \vec{W}^*$ and
$(M,\vec{u}) \sat [\vec{W} \gets \vec{w}](V \ne v)$, then $V$ takes on
a value other than that specified by the equations in 
$s_{\vec{W} \gets \vec{w},\vec{u}}$. 
Finally, say that $(M,\vec{u})$ \emph{respects the equations for $V$}
if, for all worlds $s$  such that 
$V$ takes on a value in $s$ other than that specified by
  the equations in $(M,\vec{u})$, we have $s \not\succeq s_{\vec{u}}$
  (where $\succeq$ is the preorder on worlds in $M$).  

Recall from the proof of Theorem~\ref{pro:stability}  that to show
that $A=1$ is a cause of $B=1$ in $(M_{2n+1},u_1)$, we considered
a witness world where $X_{n+1} = 0$ and $A=0$.  Once we take normality into
account,  if we require that the normality ordering in $M_{2n+1}$ be
such that $(M_{2n+1},u_1)$ respects that equations for
$X_{n+1}$, a world where $X_{n+1} = 0$ is less normal than $s_{\vec{u}}$,
so cannot be used to satisfy AC2(a).
As the following theorem shows, this observation generalizes.

\thm\label{thm:stability} If $M$ and $M'$ are extended
causal models such that (a) $M'$ is a conservative extension of $M$,
(b) $\vec{X} = \vec{x}$ is not a cause of $\phi$ in
$(M,\vec{u})$, and (c) $(M',\vec{u})$ respects the equations for all
the endogenous variables that are in $M'$ but not in $M$,
then either 
$\vec{X} = \vec{x}$ is not 
a cause of $\phi$ in $(M',\vec{u})$ 
or there is a strict subset 
$\vec{X}_1$ of 
$\vec{X}$ such that $\vec{X}_1 = \vec{x}_1$ is a cause of $\phi$ in
$(M,\vec{u})$, where
$\vec{x}_1$ is the restriction of $\vec{x}$ to the variables in $\vec{X}_1$.
\ethm
\prf  Suppose that the assumptions of the theorem hold
and that $\vec{X} = \vec{x}$ is a cause
of $\phi$ in $(M',\vec{u})$ with 
witness $(\vec{W},\vec{w},\vec{x}')$.  
I show that there is a strict subset 
$\vec{X}_1$ of 
$\vec{X}$ such that $\vec{X}_1 = \vec{x}_1$ is a cause of $\phi$ in
$(M,\vec{u})$, where
$\vec{x}_1$ is the restriction of $\vec{x}$ to the variables in $\vec{X}_1$.

Let $\V$ be the set of
endogenous variables in $M$, 
let $\vec{W}_1 = \vec{W} \inter \V$, 
let $\vec{Z}_1 = \V - \vec{W}$, and
let $\vec{w}_1$ be the restriction of $\vec{w}$ to the variables
in $\vec{W}_1$.  Since $\vec{X}= \vec{x}$ is not a cause of $\phi$ in
$(M,\vec{u})$, it is certainly not a cause with witness
$(\vec{W}_1,\vec{w}_1,\vec{x}')$. 
Thus, 
either (i) $(M,\vec{u}) \sat \vec{X}\ne \vec{x} \lor \neg\phi$ (i.e.,
AC1 is violated); (ii)  
$(M,\vec{u}) \sat [\vec{X} \gets \vec{x}', \vec{W}_1 \gets\vec{w}_1]
\phi$ (i.e., AC2(a) 
is violated), (iii) there exist subsets $\vec{W}'_1$ of $\vec{W}_1$ and
$\vec{Z}'_1$ of $\vec{Z}_1$ such that if $(M,\vec{u}) \sat \vec{Z}'_1 =
\vec{z}_1$ (i.e., $\vec{z}_1$ gives the actual values of the
variables in $\vec{Z}'_1$), then $(M,\vec{u}) \not\sat [\vec{X} \gets
  \vec{x},  \vec{W}_1' \gets \vec{w}_1, \vec{Z}_1' \gets \vec{z}_1]
\neg \phi$ (i.e., AC2(b) is 
violated), (iv) 
$s_{\vec{X} = \vec{x}', \vec{W}_1 = \vec{w}_1,\vec{u}}
\not\succeq s_{\vec{u}}$ (i.e., 
the normality condition in AC2(a$^+$) is violated), 
or (v) there is a strict subset $\vec{X}_1$ of
$\vec{X}$ such that $\vec{X}_1 = \vec{x}_1$ is a cause of $\phi$ in
$(M,\vec{u})$, where $\vec{x}_1$ is the restriction of $\vec{x}$ to the
variables in $\vec{X}_1$ (i.e., AC3 is violated). 
I now show that none of (i)--(iv) can hold, which suffices to prove
the result.

Since $M'$ is a conservative extension of $M$, 
by Lemma~\ref{lem:conservative}, 
if (i) or (iii) holds, then the same statement holds with $M$
replaced by $M'$, showing that $\vec{X} = \vec{x}$ is not a cause of
$\phi$ in $(M',\vec{u})$ with witness $(\vec{W},\vec{w},\vec{x}')$,
contradicting our assumption.   
If (ii) holds, it is still consistent that
AC2(a) holds in $M'$ with witness $(\vec{W},\vec{w},\vec{x}')$.  However if,
for each variable $V \in \vec{W} - \vec{W}_1$, if $v$ is the value of
$V$ in $\vec{w}$ and we have 
$(M',\vec{u}) \sat [\vec{X} \gets \vec{x}', \vec{W}_1 \gets \vec{w}](V 
= v)$, then $(M',\vec{u}) \sat [\vec{X} \gets \vec{x}', \vec{W} \gets\vec{w}]
\phi$, and AC2(a) also fails in $(M',\vec{u})$.  
On the other hand, if $(M',\vec{u}) \sat [\vec{X} \gets \vec{x}',
  \vec{W}_1 \gets \vec{w}](V \ne v)$ for some $V \in \vec{W} -
\vec{W}_1$, then, in the world $s_{\vec{X} = \vec{x}', \vec{W} \gets
  \vec{w},\vec{u}}$, the variable 
 $V$ takes on a value other than that specified by the
equations in $(M',\vec{u})$.  Since, by assumption, $(M',\vec{u})$ respects
the equations 
for $V$,  we have $s_{\vec{X} = \vec{x}', \vec{W}
  \gets \vec{w},\vec{u}} \not\succeq' s_{\vec{u}}$, contradicting the
assumption that $\vec{X} = \vec{x}$ is a cause of $\phi$ in
$(M',\vec{u})$ with witness $(\vec{W},\vec{w},\vec{x}')$.  Either way,
if (ii) holds, we get a contradiction.  Finally, if (iv) holds, by CE,
we must have $s_{\vec{X} = \vec{x}', \vec{W}_1 = \vec{w}_1,\vec{u}}
\not\succeq' s_{\vec{u}}$.  Moreover, as we observed in the argument
for (ii), we must have $(M',\vec{u}) \sat [\vec{X} \gets \vec{x}',
  \vec{W}_1 \gets \vec{w}](V = v)$ for each variable $V \in \vec{W} -
\vec{W}_1$, where $v$ is the value of $V$ in $\vec{w}$, or else we get
a contradiction to $\vec{X} = \vec{x}$ being a cause of $\phi$ in
$(M',\vec{u})$ with witness $(\vec{W},\vec{w},\vec{x}')$.  But this
means that $s_{\vec{X} = \vec{x}', \vec{W}_1 = \vec{w}_1,\vec{u}} = 
s_{\vec{X} = \vec{x}', \vec{W} = \vec{w},\vec{u}}$, so 
$s_{\vec{X} = \vec{x}', \vec{W} = \vec{w},\vec{u}}
\not\succeq' s_{\vec{u}}$, and again we get
a contradiction to $\vec{X} = \vec{x}$ being a cause of $\phi$ in
$(M',\vec{u})$ with witness $(\vec{W},\vec{w},\vec{x}')$.
\eprf

We immediately get that single-variable non-causality is stable.

\cor\label{cor:stability} If (a) $X = x$ is not a cause of $\phi$ in
$(M,\vec{u})$, (b) $M'$ is a conservative extension of $M$, 
and (c) 
$(M',\vec{u})$ respects the equations for all
the endogenous variables that are in $M'$ but not in $M$,
then $X = x$ is not a cause of $\phi$ in $(M',\vec{u})$.
\ecor

\commentout{
\prf  Suppose that the conditions of the theorem hold and, by way of
contradiction, that $X=x$ is a cause of $\phi$ in $(M',\vec{u})$
according to the
extended definition of causality.  Then there must
be a witness $(\vec{W},\vec{w},x')$ such that 
$s_{X \gets x', \vec{W} \gets\vec{w},\vec{u}} \succeq'
s_{\vec{u}}$.  By Theorem~\ref{thm:stability}, there
must be some variable $V \in \vec{W}$ such that $V$ is not a
variable in $M$ and $(M',\vec{u}) \sat [X \gets x', \vec{W}
  \gets \vec{w}](V \ne v)$.   Let $\V'$ be the set of endogenous
variables in $M'$, let $\vec{W}_1 = \V' - (\vec{W} \union
\{X,V\})$, and suppose that $(M',\vec{u}) \sat [X \gets x', \vec{W}
  \gets \vec{w}](\vec{W}_1 = \vec{w}_1)$.  Thus, $\vec{w}_1$ gives the
values of the variables in $\vec{W}_1$ in 
$s_{X \gets x', \vec{W} \gets\vec{w},\vec{u}}$.  It is immediate that
$(M',\vec{u}) \sat [X \gets x', \vec{W} \gets \vec{w}, \vec{W}_1 \gets
  \vec{w}_1](V \ne v)$, so in world $s_{X \gets x', \vec{W}
  \gets\vec{w},\vec{u}}$, $V$ takes on a value other than that
  specified by the equations in $(M,\vec{u})$.  But then, by
  assumption, $s_{\vec{u}} \succ^{M'} s_{X \gets x', \vec{W}
    \gets\vec{w},\vec{u}}$, and we have a contradiction.
\eprf
}

While these results shows that we get stability of causality, it comes
at a price:
the assumption that the normality ordering respects the equations for
a variable relative to a context $\vec{u}$ is clearly quite a strong
one.  Although it may seem reasonable to require that it be abnormal
for the new variables not to respect the equations in $\vec{u}$, 
recall that the
normality ordering is placed on worlds, which are complete
assignments to the endogenous variables, not on complete
assignments to both endogenous and exogenous variables.  Put another
way, in general, the normality ordering does not take the context into
account.  To see why this is important, note that in almost all of our
examples of causality in a context $\vec{u}$ in a model $M$, the
witness does not respect the equations of $\vec{u}$.  For example,  to show
that Suzy's throw is a cause of the bottle shattering in the context $u$
where both Suzy and Billy throw rocks, we consider a
witness world where neither Suzy nor Billy throw.  This world clearly does not
respect the equations of $u$, where Suzy and Billy do throw
rocks.  Nevertheless, if we ignore the context, it does not seem so
abnormal that neither Suzy nor Billy throw rocks.  

Thus, saying that the normality ordering respects the equations
for a variable $V$ relative to $\vec{u}$ is really saying that, as far
as $V$ is concerned, what happens in $\vec{u}$ is really the normal
situation.  In the assassin example used to prove
Theorem~\ref{pro:stability}, it might be better to think of the variable
$A_n$ as being three-valued: $A_n=0$ if assassin $\#n$ is present and
puts in poison, $A_n=1$ if assassin $\#n$ is present and does not 
put in poison, and $A_n=2$ if assassin $\#n$ is not present.  Clearly
the normal value is $A_n = 2$.  Take $u$ to be the context
where, in model $M_{2n+1}$, $A_n = 2$.  While the 
potential presence a number of assassins makes bodyguard putting in
antidote (part of) a cause in $(M_{2n+1},u)$, it is no longer part of
a cause once we take normality into account.  Moreover, here it does
seem reasonable to say that violating the equations for $A_n$ relative
to $u$ is abnormal.  

These observations suggest why, in general, although the assumption
that respects the equations for
the variables in $\V'-\V$ relative to the context $\vec{u}$ is a
strong one, it may not be unreasonable in practice.  Typically, the
variables that we do not mention take on their expected values, and
thus are not even noticed.

The requirement that we are talking about single-variable causality 
in Corollary~\ref{cor:stability} has some bite, but not much.  
Stability of non-causality does not hold in general, even with the
abnormality assumption, as
Example~\ref{xam:non-causal} below shows.  
However, I can show that there can be at most
one change from non-causality to causality.  It follows that we cannot
get an infinite sequence of causal models, each one a conservative
extension of the one before, where the answer to the question ``Is
$\vec{X} = \vec{x}$ a cause of $\phi$?'' alternates from ``Yes'' to
``No'' and back again under reasonable (ab)normality assumptions. 
Indeed, as is shown in the following corollary,
we cannot even get such a sequence of length 3.  

\cor\label{cor:stability2} 
If 
(a) $M_2$ is a conservative extension of $M_1$, (b) 
$M_3$ is a conservative extension of $M_2$, (c) 
$\vec{X} = \vec{x}$ is a cause of $\phi$ in 
$(M_1,\vec{u})$ and $(M_3,\vec{u})$, 
(d) $(M_2,\vec{u})$ respects the
equations  for all endogenous variables in $M_2$ not in $M_1$, and (e) 
and (e) $(M_3,\vec{u})$ respects the 
equations  for all endogenous variables in $M_3$ not in $M_2$,
then $\vec{X} = \vec{x}$ is also a cause of $\phi$ in $(M_2,\vec{u})$.
\ecor

\prf Suppose, by way of contradiction, that there is a sequence
$M_1$, $M_2$, and $M_3$ of models and a context $\vec{u}$ satisfying
the conditions of the theorem, but $\vec{X}  = \vec{x}$ is not a cause
of $\phi$ in $(M_2,\vec{u})$. 
 By Theorem~\ref{thm:stability},
there must be a strict subset $\vec{X}_1$ of $\vec{X}$ such that 
$\vec{X}_1 = \vec{x}_1$ is a cause of $\phi$ in $(M_2,\vec{u})$,
where
$\vec{x}_1$ is the restriction of $\vec{x}$ to the variables in
$\vec{X}_1$. But $\vec{X}_1 = \vec{x}_1$ cannot be a cause of $\phi$ in
$(M_1,\vec{u})$, for then, by AC3, $\vec{X} = \vec{x}$ would not be
a cause of $\phi$ in $(M_1,\vec{u})$.  By Theorem~\ref{thm:stability}
again, there must be a
strict subset $\vec{X}_2$ of $\vec{X}_1$ such that $\vec{X}_2 = \vec{x}_2$
s a cause of $\phi$ in $(M_1,\vec{u})$, where $\vec{x}_2$ is the
restriction of $\vec{x}$ to $\vec{X}_2$.    But then, by AC3, $\vec{X} =
\vec{x}$ cannot be a cause of $\phi$ in $(M_1,\vec{u})$, giving us the
desired contradiction. 
\eprf


The following example, which is a variant of the example in \cite{Hal39}
showing that a cause may involve more than one conjunct, 
shows that Corollary~\ref{cor:stability2} is the
best that we can hope for.  It is possible for a non-cause to become a
cause if it has more than one conjunct.  

\xam\label{xam:non-causal}  
$A$ votes for a candidate.  $A$'s vote is
recorded in two optical scanners $B$ and $C$.   $D$ collects the
output of the scanners.  The candidate wins (i.e., $\WIN =1$) if any of
$A$, $B$, or $D$ is 1.  The value of $A$ is determined by the
exogenous variable.  The following structural equations characterize the
the remaining variables:  $B=A$, $C=A$, $D = B\land C$, $\WIN = A \lor
B \lor D$.  
Call the resulting causal model $M$.  
In the actual context $u$, $A=1$, so $B=C=D=\WIN=1$.
Assume that all worlds in $M$ are equally normal. 

I claim that $B = 1$ is a cause of $\WIN=1$ in $(M,u)$.
To see this, take $\vec{W} = \{A\}$.  
Consider the contingency where $A=0$.  Clearly if $B =
0$, then $\WIN=0$, while if $B=1$, $\WIN=1$.  It is easy to check that
AC2 holds.  Moreover, since $B = 1$ is a cause of $\WIN = 1$ in
$(M,u)$, by AC3, $B = 1 \land C = 1$ cannot be a cause of $\WIN = 1$
in $(M,u)$. 

Now consider the model $M'$ that is just like $M$, except that there is
one more exogenous variable $D'$, where $D' = B \land \neg A$.  The
equation for $\WIN$ now becomes $\WIN = A \lor D' \lor D$.  All the
other equations in $M'$ are the same as those in $M$.
Roughly speaking, $D'$ acts like $\BH$ in the rock-throwing example.  
Define the normality ordering in $M'$ so that it respects the
equations for $D'$ in $(M',u)$: all worlds where $D'=B \land \neg A$
are equally normal, all worlds where $D' \ne B \land \neg A$ are also equally
normal, but less normal than worlds where $D' = B \land \neg A$.

It is easy to see that $M'$ is a conservative extension of $M$.  Since
$D'$ does not affect any variable but $\WIN$ and all the equations
except that for $\WIN$ are unchanged, it suffices to show that for all
settings of the variables other than $D'$ and $\WIN$, $\WIN$ has the
same value in context $u$ in both $M$ and $M'$.  Clearly if $A=1$
or $D=1$, then $\WIN = 1$ in both $M$ and $M'$.  So suppose that we set
$A=D=0$.  Now if $B = 1$, then $D'=1$ (since $A=0$), so again $\WIN=1$
in both $M$ and $M'$.  On the other hand, if $B = 0$, then
$D'=0$, so $\WIN = 0$ in both $M$ and $M'$.  Condition CE clearly
holds as well.

Finally, as I now show, $B=1 \land C=1$ is a cause of $\WIN=1$
in $(M',u)$.  To see this, 
first observe that AC1 clearly holds.  For AC2, let $\vec{W} =
\{A\}$ (so $\vec{Z} = \{B,C,D,D',\WIN\}$) and take $w = 0$ (so we are
considering 
the contingency where $A=0$).  Clearly, $(M,\vec{u}) \sat [A \gets 0, B
\gets 0, C \gets 0](\WIN=0)$, so AC2(a) holds, and $(M,\vec{u}) \sat
[A=0,B=1,C=1](\WIN=1) \gets 1$.  Moreover, 
$(M,\vec{u}) \sat
[B \gets 1,C \gets1](\WIN=1)$, and $\WIN=1$ continues to hold even if $D$
is set to 1 and/or $D'$ is set to 0 (their values in $(M,u)$).  Thus,
AC2(b) holds.  

It remains to show that AC3 holds and, in particular, that neither
$B=1$ nor $C=1$ is a cause of $\WIN=1$ in $(M',u)$.  The argument is the
same for both $B=1$ and $C=1$, so I just show it for $B=1$.   Roughly
speaking, $B = 1$ is not a cause of
$\WIN=1$ for essentially the same reason that $\BT=1$ is not a cause of
$\BS=1$.  For suppose that $B=1$ were a cause.  Then we would have to have
$A \in \vec{W}$, and we would need to consider the contingency where
$A=0$ (for otherwise $\WIN=1$ no matter how we set $B$).  Now we need to
consider two cases: $D' \in \vec{W}$ and $D' \in \vec{Z}$.  If $D' \in
\vec{W}$, then if we consider the 
contingency where $D' = 0$, we have $(M',u) \sat [A \gets 0, B \gets
1, D' \gets 0](\WIN = 0)$, so AC2(b) fails (no matter whether $C$
and $D$ are in $\vec{W}$ or $\vec{Z}$).  And if we consider the
contingency where $D' = 1$, then AC2(a) fails, since 
$(M',u) \sat [A \gets 0, B \gets
0, D' \gets 1](\WIN = 1)$.  Now if $D' \in \vec{Z}$, note that $(M,u)
\sat D' = 0$.  Moreover, as we have observed, $(M,u) \sat
[A=0,B=1,D'=0](\WIN=0)$, so again AC2(b) fails (no matter whether $C$
or $D$ are in $\vec{W}$ or $\vec{Z}$).  Thus, $B=1$ is not a cause of
$\WIN=1$ in $(M',u)$.  
%
Thus, $B=1 \land C=1$ goes from not being a cause of $\WIN=1$ in $(M,u)$ to
being a cause of $\WIN=1$ in $(M',u)$.  

Now consider the model $M''$ which is just like $M'$ except that it
has one additional variable $D''$, where $D'' = D \land \neg A$ and
the equation for $\WIN$ becomes $\WIN = A \lor D' \lor D''$.  All the
other equations in $M''$ are the same as those in $M'$.  
Define the normality ordering in $M''$ so that it respects the
equations for both $D'$ and $D''$ in $(M'',u)$.

It is easy to check that $M''$ is a conservative extension of $M'$.  
Since $D''$ does not affect any variable but $\WIN$ and all the equations
except that for $\WIN$ are unchanged, it suffices to show that for all
settings of the variables other than $D''$ and $\WIN$, $\WIN$ has the
same value in context $u$ in both $M$ and $M'$.  Clearly if $A=1$
or $D'=1$, then $\WIN = 1$ in both $M'$ and $M''$.  And if $A=D=0$,
then $D'=1$ iff $D=1$, so again the value of $\WIN$ is the same in
$M'$ and $M''$.  Condition CE clearly holds as well.

Finally, I claim that $B=1 \land C=1$ is no longer a cause of $\WIN=1$
in $(M'',u)$.  Suppose, by way of
contradiction, that it is, with 
witness $(\vec{W}, \vec{w}, \vec{x}')$. $A=0$ must be a conjunct of
$\vec{W} = \vec{w}$.
It is easy to see that either
$D'=0$ is a conjunct of $\vec{W} = \vec{w}$ or $D' \notin \vec{W}$,
and similarly for 
$D''$.  Since $D'=D''=0$ in the context $u$, and 
$(M'',u) \sat [A \gets 0, D' \gets 0, D'' \gets 0](\WIN = 0)$, it
easily follows that AC2(b) does not hold, no matter whether $D'$
and $D''$ are in $\vec{W}$.  

Thus, $B=1 \land C=1$ goes from not being a cause of
$\WIN=1$ in $(M,u)$ to being a cause of $\WIN=1$ in $(M',u)$ to not
being a cause of of $\WIN=1$ in $(M'',u)$. 
\exam


\section{Conclusions}\label{sec:conclusions}

This paper has demonstrated the HP definition
of causality is remarkably resilient, but it emphasizes how sensitive
the ascription of causality can be to the choice of model.
The focus has been on showing that the choice of variables is a powerful
modeling tool.  But it is one that can be abused.  One lesson that comes
out clearly is the need to have 
variables that describe the mechanism of causality, particularly if
there is more than one mechanism.
However, this is hardly a general recipe.  Rather, it is a heuristic for
constructing  a ``good'' model.  As Halpern and Hitchcock
\citeyear{HH10} point out,
constructing a good model is still more of an art than a science.

The importance of the choice of 
variables to the ascription of causality leads to an obvious
question: to what extent is the choice of variables determined by the
story.  Certainly some variables are explicit in a causal story.
If we talk about Suzy and Billy throwing rocks at a bottle,
which shatters, it seems pretty clear that a formal model needs to
have variables that talk about Suzy and Billy throwing rocks, and the
bottle shattering.  
Furthermore, if the story says that Suzy's rock hits first, it also 
seems clear that we need variables in the formal model to capture the
fact that Suzy's rock hit first.  Unfortunately, there is more than
one way to capture this fact using variables.  Here I used the
variables $\SH$ and $\BH$, as was done in \cite{HP01b}.  But in
\cite{HP01b}, another model was also presented, where the 
there are time-indexed variables (e.g., a 
family of variables $\BS_k$ for ``bottle shatters at time $k$'').
In the model with time-indexed variables it is still the case that
Suzy's throw is a cause of the bottle shattering and Billy's throw is
not.  The point here is that the story does not make explicit which
variables should be used.  While a modeler must ultimately justify
whatever variables are used in terms of how well they 
capture the intent of the story, there is clearly a lot left to the
modeler's judgment here.  (A similar point is made in \cite{HH10}.) 

A second lesson of this paper is that there is an interplay between
the choice of variables and normality considerations.  Moreover,
normality considerations can 
play quite an important role in dealing with issues regarding the
stability of causality and non-causality.
There are doubtless other lessons that
will be learned as we get more experience with causal modeling.
Structural models are a powerful tool for modeling causality, but they
have to be handled with care!

\fullv{\bibliographystyle{chicagor}}
\shortv{\bibliographystyle{aaai}}
\bibliography{z,joe,refs}

\end{document}